\documentclass[10pt,twocolumn,letterpaper]{article}

\usepackage{cvpr}              
\usepackage{adjustbox}
\usepackage{xcolor}

\definecolor{cvprblue}{rgb}{0.21,0.49,0.74}
\usepackage[pagebackref,breaklinks,colorlinks,allcolors=cvprblue]{hyperref}

\hypersetup{
    colorlinks=true,
    linkcolor=blue,
    filecolor=magenta,      
    urlcolor=black,
}
\usepackage[font=small, skip=3pt]{caption}
\usepackage{color, colortbl}
\usepackage{booktabs}
\usepackage{arydshln}
\usepackage{multirow}

\def\paperID{1830} 
\def\confName{CVPR}
\def\confYear{2025}

\newcommand\norm[1]{\left\lVert#1\right\rVert}

\newcommand{\hb}{{\boldsymbol h}}
\newcommand{\xb}{{\boldsymbol x}}

\newcommand{\yb}{{\boldsymbol y}}

\newcommand{\zb}{{\boldsymbol z}}

\newcommand{\fb}{{\boldsymbol f}}
\newcommand{\epsilonb}{{\boldsymbol \epsilon}}

\newcommand{\Eb}{{\mathbb E}}

\usepackage{amsthm}

\newcommand{\Nc}{{\mathcal N}}

\usepackage{xspace}
\newcommand{\method}{Track4Gen\xspace}
\newcommand{\diffusionloss}{diffusion loss\xspace}
 
\newcommand\Mark[1]{\textsuperscript#1}

\definecolor{bostonuniversityred}{rgb}{0.8, 0.0, 0.0}
\definecolor{cadmiumgreen}{rgb}{0.0, 0.42, 0.24}
\definecolor{calpolypomonagreen}{rgb}{0.12, 0.3, 0.17}
\hypersetup{
    urlcolor=cadmiumgreen,
}

\title{Track4Gen: Teaching Video Diffusion Models to Track Points \\ Improves Video Generation}

\author{
    Hyeonho Jeong\Mark{1}\Mark{,}\Mark{2}\Mark{,}\Mark{*} \quad
    Chun-Hao P. Huang\Mark{1} \quad
    Jong Chul Ye\Mark{2} \quad
    Niloy J. Mitra\Mark{1}\Mark{,}\Mark{3} \quad
    Duygu Ceylan\Mark{1} \\[0.09in]
    \Mark{1}Adobe Research \qquad
    \Mark{2}KAIST \qquad
    \Mark{3}University College London \\
}

\begin{document}
\maketitle
\begin{abstract}

While recent foundational video generators produce visually rich output, they still struggle with appearance drift, where objects gradually degrade or change inconsistently across frames, breaking visual coherence. 
We hypothesize that this is because there is no explicit supervision in terms of spatial tracking at the feature level. 
We propose \method, a spatially aware video generator that combines video \diffusionloss with point tracking across frames, providing enhanced spatial supervision on the diffusion features. 
\method merges the video generation and point tracking tasks into a single network by making minimal changes to existing video generation architectures. 
Using Stable Video Diffusion~\cite{blattmann2023stable} as a backbone, 
\method demonstrates that it is possible to unify video generation and point tracking, which are typically handled as separate tasks. 
Our extensive evaluations show that \method effectively reduces appearance drift, resulting in temporally stable and visually coherent video generation.
Project page: \href{https://hyeonho99.github.io/track4gen}{hyeonho99.github.io/track4gen}.

\let\thefootnote\relax\footnotetext{$^*$Work done during internship at Adobe.}
\end{abstract}    
\section{Introduction}
\label{sec:intro}

Diffusion-based video generators~\cite{blattmann2023stable, videoworldsimulators2024, polyak2024movie} are making rapid strides in creating temporally consistent and visually rich video content.
This progress marks a significant shift, as the unification of generation and control has the potential to transform the traditional workflow of first capturing and then digitally editing video.

\begin{figure}[t!]
    \centering
    \includegraphics[width=\columnwidth]{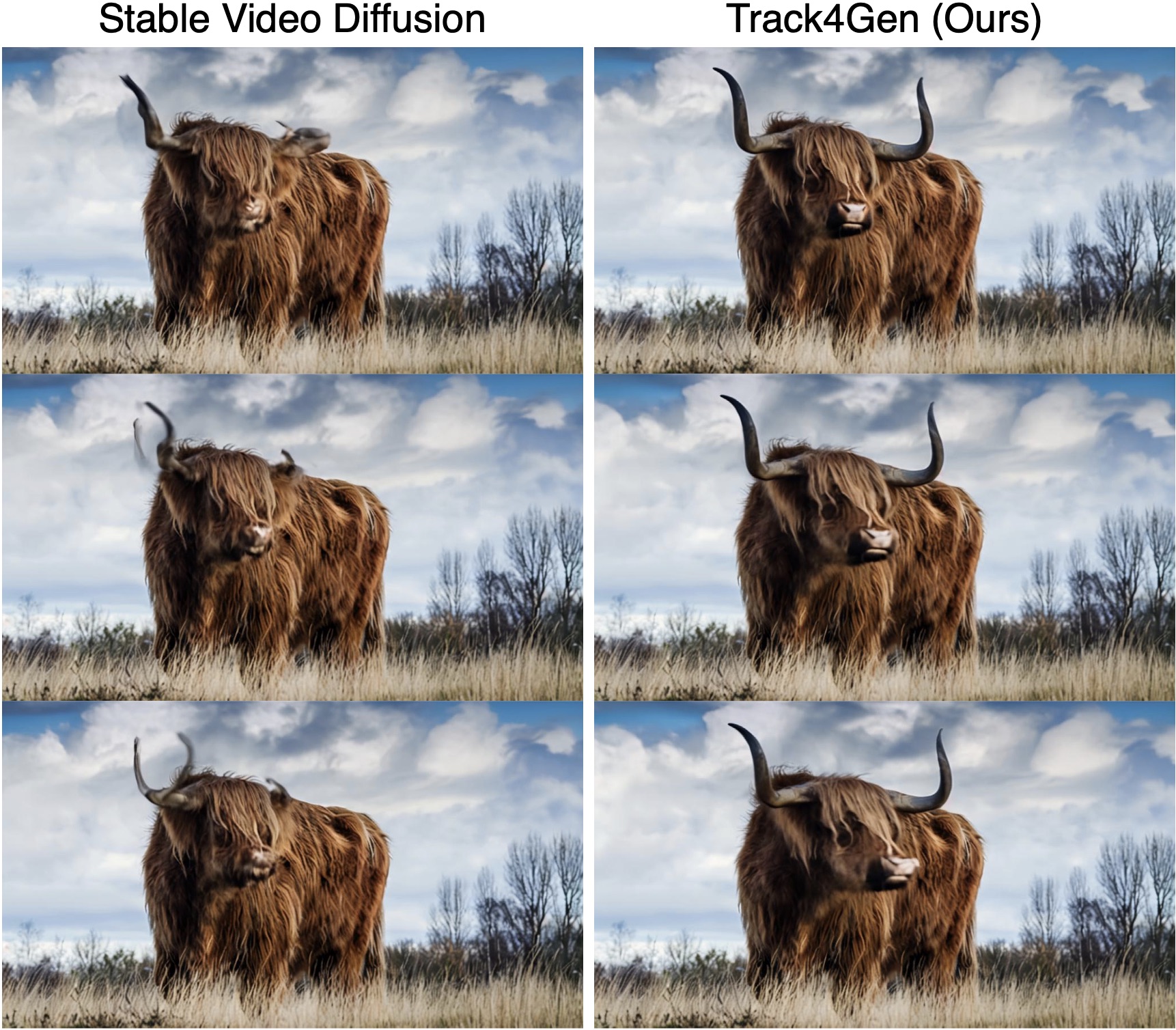}
    \caption{
    \textbf{Motivation.}
    Videos generated by Stable Video Diffusion \cite{blattmann2023stable} suffer from appearance drift, while those from our method, Track4Gen, are free from such appearance inconsistency issues.
    }
    \label{fig:motivation}
\end{figure}

Despite impressive capabilities, video generators often suffer from \textit{appearance drift}, where visual elements gradually change, mutate, or degrade over time, causing inconsistencies in the objects. 
For example, in Fig. \ref{fig:motivation}, we observe the horns of the cow distorting and morphing unrealistically over time, breaking the plausibility of the generated content. 
This is in striking contrast to humans, who develop a sense of \textit{appearance constancy} as early as infancy through observation and interaction with the world ~\cite{yang2015pre}.

Unfortunately, appearance drift remains a persistent issue in current video models, even with increased training data and more advanced architectures.
We speculate that this limitation arises from supervision being based solely on video \diffusionloss (i.e., denoising score matching \cite{vincent2011connection}) in the pixel/latent space, without explicit spatial awareness guidance in the feature space.
Hence, in this paper, we ask \textit{if and how we can empower video diffusion models with appearance constancy by providing additional supervision}.   

We present \textit{\method} as a spatially aware video generator that receives supervision both in terms of the original diffusion-based objective as well as (dense) point correspondence across frames, which we refer to as  \textit{tracks}.
We demonstrate that it is possible to provide such track-level supervision in the diffusion feature space by making minimal architecture changes. Our generated videos do not suffer from degradation of video quality (according to the usual video generation metrics), while being significantly more spatially coherent as the highlight cow in Fig. \ref{fig:motivation}.

We train \method using the latest Stable Video Diffusion~\cite{blattmann2023stable} as the backbone and evaluate on the publicly available VBench dataset~\cite{huang2024vbench}. We report significant improvement in terms of appearance constancy of subjects, both in quantitative and qualitative (i.e., via user studies) evaluations. In summary, we demonstrate that it is possible to upgrade existing video generators, by supervising them with additional correspondence tracking loss, to produce videos without significant appearance drifts, a problem commonly encountered in diffusion-based video generators.

\section{Related Work}
\label{sec:related work}

\paragraph{Diffusion-based video generation.}
Building on the success of diffusion models in image synthesis \cite{dhariwal2021diffusion, rombach2022high}, diffusion-based video generators have seen significant advancements~\cite{ho2022video, blattmann2023stable, videoworldsimulators2024, polyak2024movie}. 
A commonly adopted approach is to extend text-to-image models to the video domain by incorporating temporal layers to facilitate interactions across video frames~\cite{blattmann2023align, singer2022make, guo2023animatediff}.
While some works have adopted cascaded approaches to produce both spatially and temporally high-resolution videos \cite{singer2022make, ho2022imagen, zhang2023show, wang2023lavie, zhang2023i2vgen, polyak2024movie}, others have utilized lower-dimensional latent space modeling to reduce computational demands \cite{he2022latent, blattmann2023align, chen2023videocrafter1, zhou2022magicvideo}. 
We build on top of one such approach, Stable Video Diffusion (SVD, \cite{blattmann2023stable}), which introduces a latent image-to-video diffusion model trained on a large-scale and curated video data.

With advances in generation, systematic evaluation of generation quality has become crucial.
Traditionally, metrics such as Fr\'echet Inception Distance (FID, \cite{heusel2017gans}), Fr\'echet Video Distance (FVD, \cite{unterthiner2018towards}), and CLIPSIM \cite{radford2021learning} are used.
Additionally, comprehensive benchmark suites \cite{huang2024vbench, wu2024towards} have been introduced to provide a more robust evaluation aligned with human perception. Inspired by such work, we thoroughly evaluate our approach and demonstrate improved video generation quality with respect to both conventional metrics and the recent VBench metrics \cite{huang2024vbench}.

\paragraph{Foundational models as feature extractors.}
Various foundational models such as vision transformers~\cite{DosovitskiyB0WZ21} or diffusion-based generators~\cite{Rombach2021HighResolutionIS} have been utilized as feature extractors for various tasks including semantic matching~\cite{hedlin2023unsupervised,li2023sd4match,Dutt_2024_CVPR}, classification~\cite{li2023diffusion}, segmentation~\cite{xu2022odise,wang2024zeroshot}, and editing~\cite{tumanyan2023plug, geyer2023tokenflow, gu2024videoswap}. There have been efforts to boost their performance by post-processing the feature maps obtained from the pre-trained models, e.g., by upsampling~\cite{fu2024featup,suri2025lift}. In a recent effort, Yue et al.~\cite{yue2024fit3d} lift semantic per-frame features from a foundational model into a 3D Gaussian representation. They fine-tune the foundational model with such 3D-aware features resulting in improved performance in downstream tasks. Similarly, Sundaram et al.~\cite{sundaram2024doesperceptualalignmentbenefit} fine-tune state-of-the-art foundational models on human similarity judgments yielding improved representations across downstream tasks. 
In a concurrent effort, Yu et al.~\cite{yu2024repa} propose to align the internal features of an image generation model with external discriminative features~\cite{oquab2023dinov2}, which results in more effective training of the generator.

Our work also enhances the internal feature representation of a foundational generation model but with significant differences compared to previous literature. First, unlike most previous work that focus on image level foundational models, we exploit the power of recently emerging video models. Second, instead of post-processing, we enhance the spatial awareness of the intermediate features by training the generator to jointly perform an additional tracking task. 
We show that this joint training  boosts the performance of intermediate features in correspondence tracking, leading to improved video generation quality.

\begin{figure*}[!t]
    \centering
    \includegraphics[width=0.97\textwidth]
    {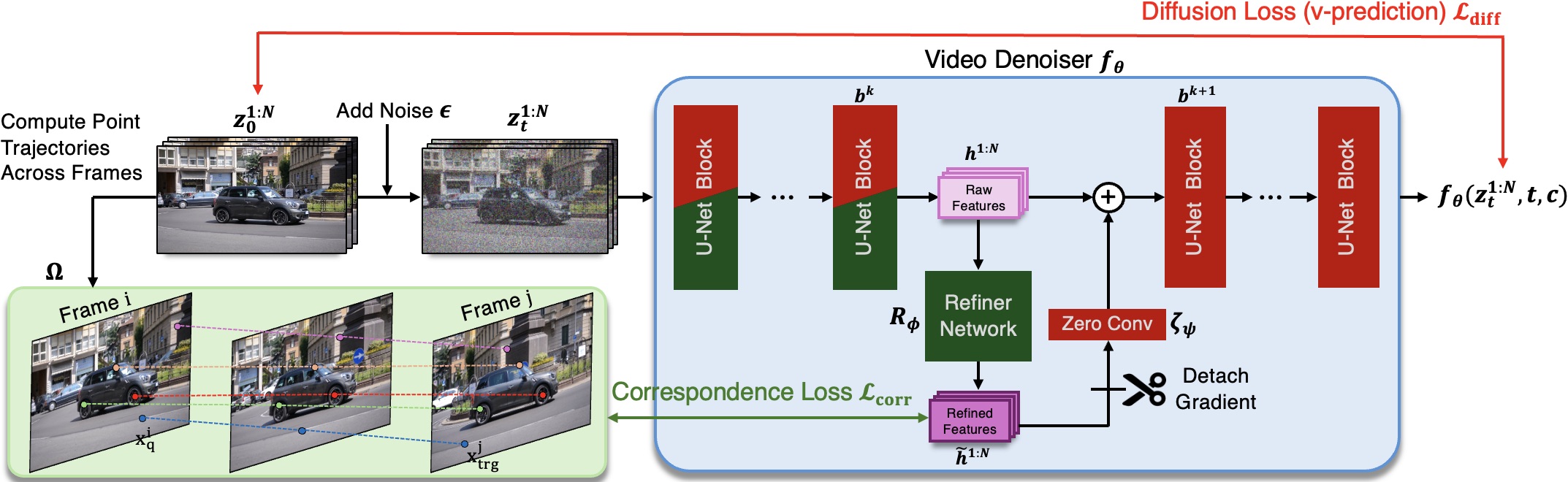}
    \vspace{2mm}
    \caption{
    \textbf{Track4Gen overview}.
    \textcolor{bostonuniversityred}{Red}-colored blocks represent layers optimized by the diffusion loss \textcolor{bostonuniversityred}{$\mathcal{L}_{\text{diff}}$}, while \textcolor{cadmiumgreen}{green} blocks are optimized by the correspondence loss \textcolor{cadmiumgreen}{$\mathcal{L}_{\text{corr}}$}. Blocks colored both \textcolor{bostonuniversityred}{red} and \textcolor{cadmiumgreen}{green} are influenced by the joint loss, $\textcolor{bostonuniversityred}{\mathcal{L}_{\text{diff}}} + \lambda \textcolor{cadmiumgreen}{\mathcal{L}_{\text{corr}}}$.
See text for details. 
    }
    \label{fig: overview}
\end{figure*}

\paragraph{Tracking any point in a video.}
The task involves following any arbitrary query point across a long video sequence. 
First introduced by PIPs \cite{harley2022particle} and later re-framed by TAP-Vid \cite{doersch2022tap}, several methods have emerged in recent years to tackle long-term point tracking. 
PIPs \cite{harley2022particle} revisits the classical particle-based representation \cite{sand2008particle} and introduces MLP-based networks that predict point tracks within an 8-frame window.
Subsequent works have improved performance by capturing longer temporal context through advanced architectures \cite{bian2023context, harley2022particle, doersch2023tapir, karaev2023cotracker}, as well as by enabling the simultaneous tracking of multiple queries \cite{karaev2023cotracker, cho2024local}. More recent training-based trackers \cite{xiao2024spatialtracker, li2024taptr, cho2024local, karaev2024cotracker3} have achieved remarkable performance by leveraging high-capacity neural networks to learn robust priors from large-scale training data.
While high-quality data is crucial for accurate tracking, manually annotating point tracks is prohibitively expensive. 
Hence, synthetic videos \cite{greff2022kubric} with automatic annotations, have become an alternative and have demonstrated effectiveness in real-world video tracking. An alternative approach is self-supervised adaptation at test time, where tracking is learned without ground-truth labels \cite{jabri2020space, wang2023tracking, tumanyan2024dino}. 
In a recent study, Aydemir et al.~\cite{aydemir2024can} evaluate the effectiveness of several image foundational model features for point tracking both in zero-shot setting as well as with supervised training using low-rank adapter layers.  
To the best of our knowledge, we are the first to exploit the features of a foundational video diffusion model for dense point tracking. 
\section{Method}
\label{sec: method}
In this section, we provide a comprehensive discussion of the \method framework. 
We begin with a concise overview of latent video diffusion models (Sec. \ref{sec: method-background}).
Next, we discuss how video diffusion features relate to temporal correspondences both for real and generated videos (Sec. \ref{sec: method-observations}).
Finally, we detail the design of \method both in terms of network architecture and the employed supervision signals (Sec. \ref{sec: method-track4gen}).
An overview is depicted in Fig. \ref{fig: overview}.

\subsection{Background: Stable Video Diffusion}
\label{sec: method-background}

Starting from random Gaussian noise, diffusion models aim to generate clean images or videos via an iterative denoising process \cite{sohl2015deep, ho2020denoising}.
This process reverses a fixed, time-dependent diffusion forward process, which gradually corrupts the data by adding Gaussian noise.
While our method is applicable to general video diffusion models, in this paper, we design our architecture based on Stable Video Diffusion (SVD), a latent video diffusion model which employs the EDM-framework \cite{karras2022elucidating}. The diffusion process operates in the lower-dimensional latent space of a pre-trained VAE \cite{kingma2013auto}, consisting of an encoder $\mathcal{E}(\cdot)$ and a decoder $\mathcal{D}(\cdot)$.

Given a clean sample $\xb_0^{1:N} \sim p_{\text{data}}(\xb)$ of an $N$-frame video sequence, the frames are first encoded into the latent space as $\zb_0^{1:N} = \mathcal{E}(\xb_0^{1:N})$.
Gaussian noise $\epsilonb \sim \mathcal{N}(0, I)$ is then added to the latents to produce the intermediate noisy latents via the forward process $\zb_t^{1:N} = \alpha_t \zb_0^{1:N} + \sigma_t \epsilonb,$
where $t$ represents the diffusion timestep, and $\alpha_t$, $\sigma_t$ are the discretized noise scheduler parameters.
The diffusion denoiser $\fb_{\theta}$ is trained by minimizing the v-prediction loss:
\begin{equation}
    \label{eq: epsilon matching}
    \min_{\theta} \Eb_{\epsilonb \sim \Nc(0, I), t \sim U[1,\textit{T}]} \big[ \norm{\fb_{\theta} (\zb_t^{1:N}, t, c) - \yb}_2^2 \big],
\end{equation}
where $\yb$ is defined as $\yb = \alpha_t \epsilonb - \sigma_t \zb_0^{1:N}$.
In the image-to-video variant of SVD, the condition $c$ refers to the CLIP image embedding \cite{radford2021learning}, replacing the typical text embeddings.
For the remainder of this paper, we will refer to Eq. \ref{eq: epsilon matching} as the \textit{video \diffusionloss} $\mathcal{L}_{\text{diff}}$.

Once trained, the diffusion model generates videos by iteratively denoising a noisy latent $\zb_T^{1:N}$ sequence sampled from pure Gaussian distribution. 
At each diffusion step, the model predicts the noise in the input latent. 
Once the clean latent $\zb_0^{1:N}$ is obtained, the decoder $\mathcal{D}$ maps it to the higher-dimensional pixel space $\xb_0^{1:N} = \mathcal{D}(\zb_0^{1:N})$.
For further details, we refer to the Appendix D of \cite{blattmann2023stable}.

\begin{figure*}[!t]
    \centering
    \includegraphics[width=0.98\textwidth]
    {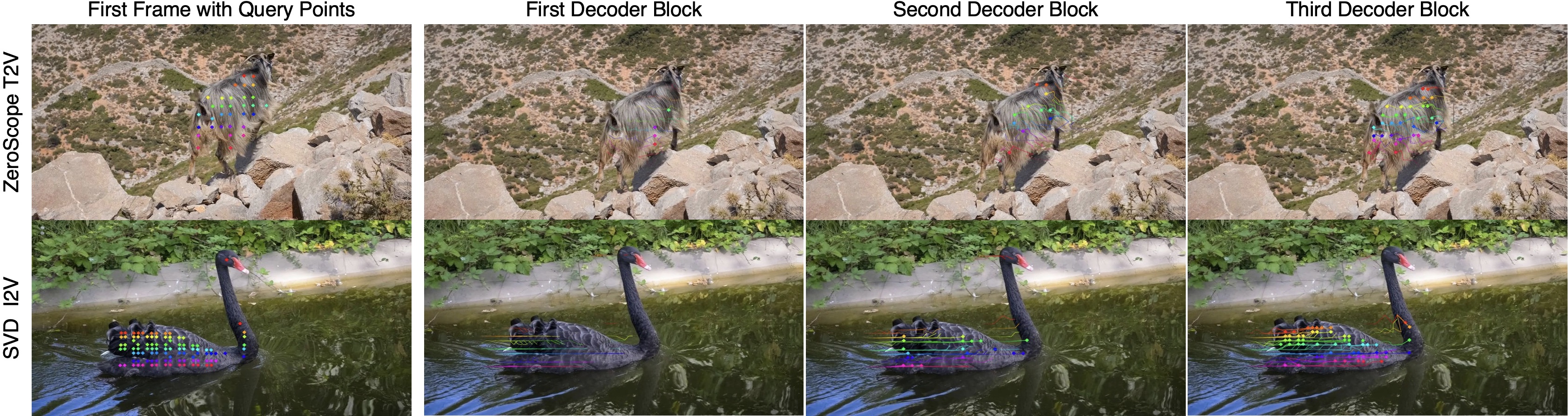}
    \caption{\textbf{Real-world video tracking using different video diffusion features}. Given color-coded query points on the first frame (\textit{Leftmost column}), we display tracked points on target frames using features from different blocks (\textit{right columns}).
    The 13th frame (\textit{first row}) and 8th frame (\textit{second row}) are shown as target frames.
    Full results are available in the supplementary and on \href{https://hyeonho99.github.io/track4gen/}{our page}.}
    \label{fig: observation-1}
\end{figure*}

\begin{figure}[b!]
    \centering
    \includegraphics[width=\columnwidth]
    {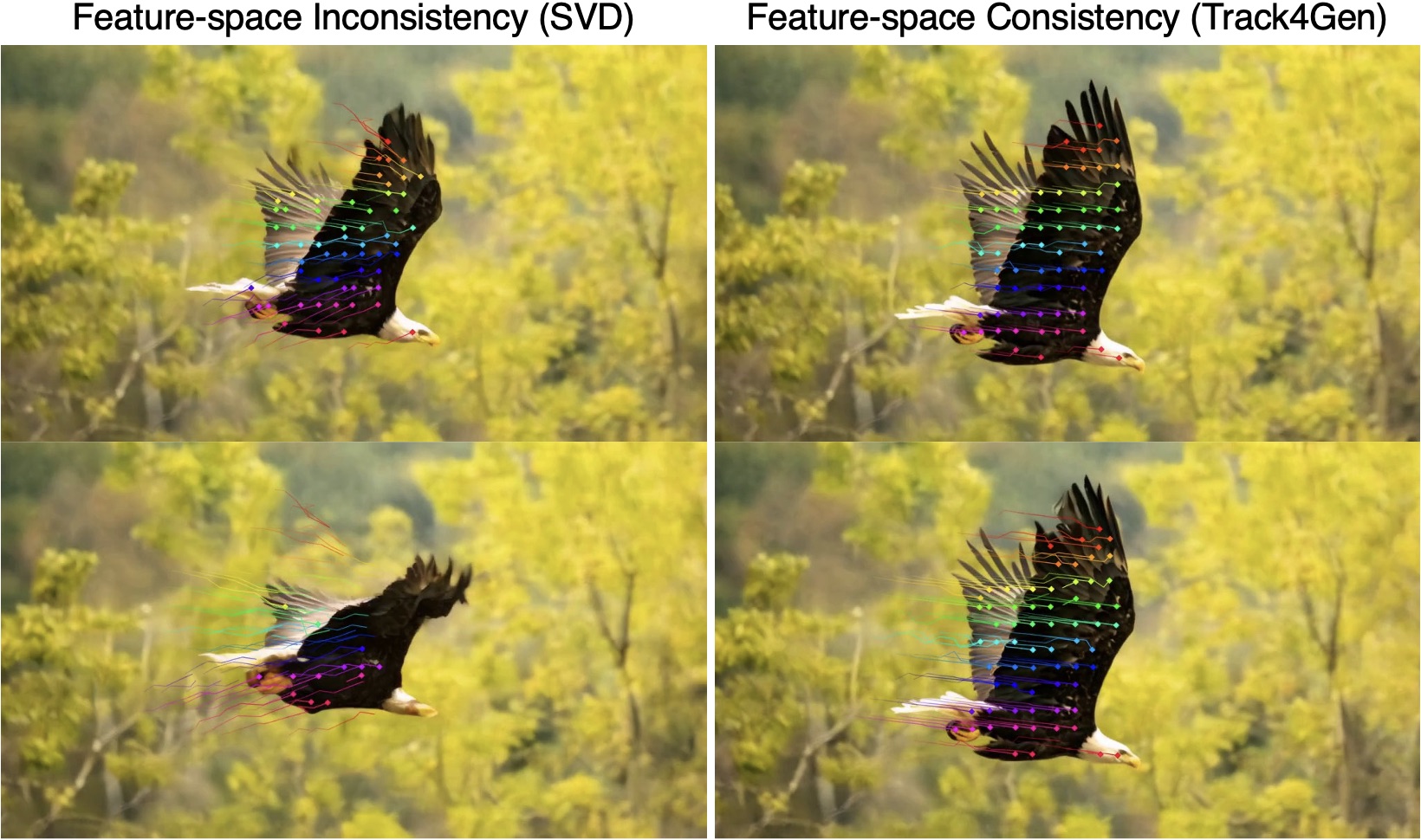}
    \caption{
    \textbf{Generated video tracking using video diffusion features.}
    Tracks based on diffusion features are annotated on the generated videos. Track4Gen generates more consistent results. 
    }
    \label{fig: observation-2}
\end{figure}

\subsection{Video Diffusion Features}
\label{sec: method-observations}
Previous studies have demonstrated that image diffusion models learn discriminative features in their hidden states that are effective for various analysis tasks and propose methods for improving the representation power of such features \cite{xiang2023denoising, chen2401deconstructing, yang2023diffusion, yu2024repa}. Similarly, we argue that while also being powerful, internal representations of pre-trained video diffusion models may not be fully temporally consistent, resulting in appearance drift in generated videos.

To better investigate this hypothesis, we first evaluate the long-term video tracking capabilities of U-Net-based video diffusion models \cite{zeroscope, zhang2023i2vgen, blattmann2023stable}. Specifically, we evaluate the effectiveness of the features from each block of the U-Net for the task of point tracking.
Given a real-world video, we add a small amount of noise and extract feature maps from each layer in each block. We perform a cosine-similarity-based nearest-neighbor search \cite{tang2023emergent, luo2024diffusion} over these feature maps for a given set of fixed query points on the first frame (we use a similarity threshold of 0.6 \cite{tumanyan2024dino} in our experiments). We also perform a similar analysis for generated videos where we extract the feature maps corresponding to diffusion steps with small amount of noise.

Based on this feature analysis, we make some important observations. Notably, regardless of the model (we analyze both Zeroscope T2V~\cite{zeroscope} and SVD I2V~\cite{blattmann2023stable}), we find out that output features from the \textit{upsampler layer of the third decoder block} consistently yield stronger temporal correspondences, as shown in Fig. \ref{fig: observation-1}. Hence, we use this block when extracting features for the remainder of our experiments. Furthermore, when we analyze generated videos and point tracks estimated based on the feature maps (as shown in Fig. \ref{fig: observation-2}), we observe that there is a correlation between \textit{tracking failures} that reveal feature-space inconsistencies and \textit{appearance drifts} that reveal pixel-space inconsistencies. Hence, we hypothesize that enriching feature consistency can help mitigate such appearance drifts. Next, we introduce \method where we accomplish this goal by supervising video diffusion models with a joint tracking loss.
%
%

\begin{figure*}[!t]
    \centering
    \includegraphics[width=\textwidth]
    {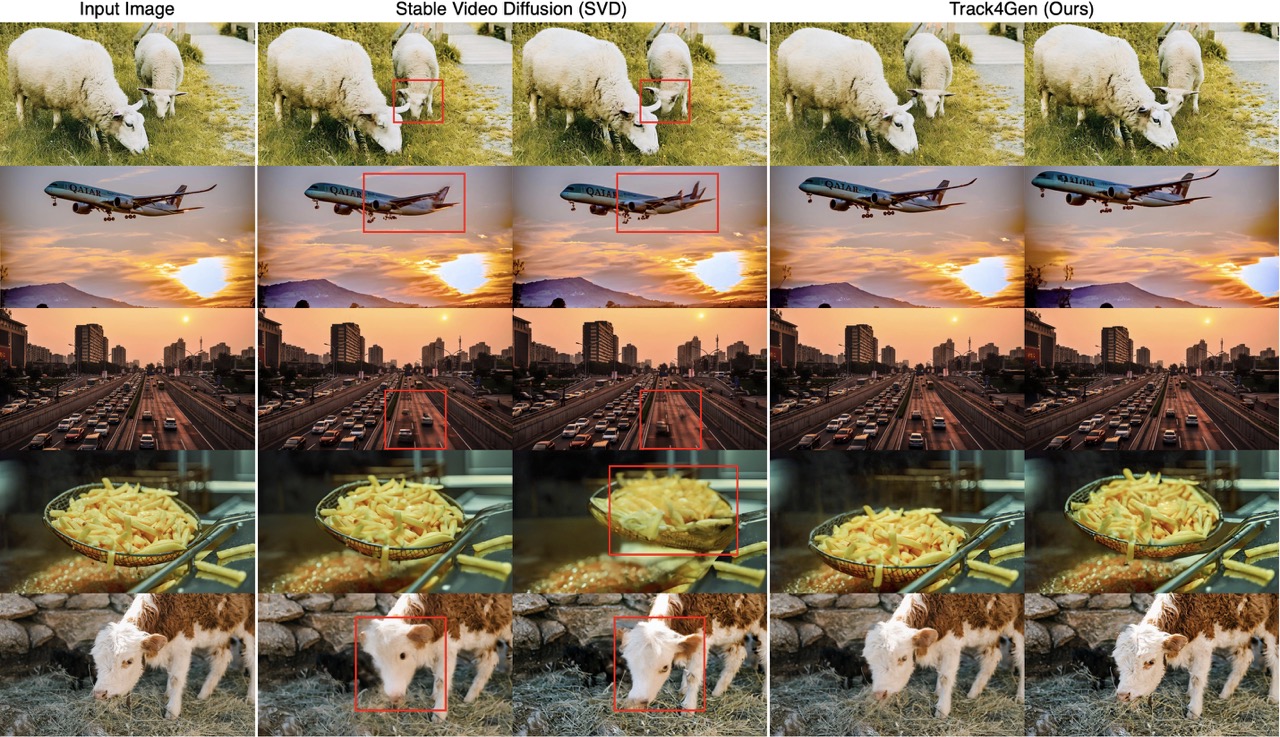}
    \caption{
    \textbf{Image-to-video generation results of the original SVD and Track4Gen.} 
    Please visit \href{https://hyeonho99.github.io/track4gen/page2.html}{our page} for full video view.
    }
    \label{fig: video-gen-qualitative}
\end{figure*}

\subsection{Track4Gen}
\label{sec: method-track4gen}
%
%

\method aims to utilize point tracking as an additional supervision signal to enhance the spatial-awareness of video diffusion features. Given that we build on top of a pre-trained video generation model, to retain the prior knowledge and avoid tampering the original features directly, we propose a novel architecture change as shown in Fig. \ref{fig: overview}. Specifically, instead of directly using the raw diffusion features for correspondence estimation, we propose a trainable \emph{refiner module} $\boldsymbol{R}_{\phi}$, which is designed to refine the raw features by projecting them into a correspondence-rich feature space. 
The refined features, which are spatially-aware, are then both used to estimate point tracks with an explicit supervision as well as feeding back to the generation backbone. We empirically find out that this design is more effective compared to fine-tuning the original model with no refinement module (see Sec.~\ref{sec: vid-gen-comparison}).

Given an $N$-frame video sequence $\xb_0^{1:N}$, its corresponding latent $\zb_0^{1:N}$, and a diffusion timestep $t$, in order to train Track4Gen we continue to utilize the standard diffusion training loss as defined in Eq. (\ref{eq: epsilon matching}), where we adopt the velocity prediction objective \cite{karras2022elucidating, salimans2022progressive, blattmann2023stable} for $\mathcal{L}_\text{diff}$.

To enable tracking supervision, we assume access to a dense set of point trajectories $\Omega = \{(\boldsymbol{\text{x}}^i, \boldsymbol{\text{x}}^j) \}$ across frames  where a point $\boldsymbol{\text{x}}^i$ in frame $i$ corresponds to a matching point $\boldsymbol{\text{x}}^j$ in frame $j$ and vice versa. Given the corresponding noisy video latent sequence $\zb_t^{1:N}$, we first extract raw diffusion features as the hidden states $\hb^{1:N} \in \mathbb{R}^{N \times H \times W \times C}$ from a specific block $b^k$ within the U-Net, where $b^k$ is set to the upsampler layer of the third decoder block (see Sec. \ref{sec: method-observations}). We then pass these features through the refiner module to obtain the refined feature map $\Tilde{\hb}^{1:N} = \boldsymbol{R}_{\phi}(\hb^{1:N})$.

We sample a query point $\text{x}_\text{q}^i$ along with its ground-truth target point $\text{x}_\text{trg}^j$ from the correspondence set $\Omega$.
Given the query point feature $\Tilde{\hb}^{i}(\text{x}_\text{q}) \in \mathbb{R}^{1 \times 1 \times C}$ and the target feature map $\Tilde{\hb}^{j} \in \mathbb{R}^{H \times W \times C}$, we calculate the cost volume $\boldsymbol{S} \in \mathbb{R}^{H \times W \times 1}$ as follows:
\begin{equation}
    \label{eq: cost-volume}
    \boldsymbol{S}(\text{p}) = \text{cos-sim}(\Tilde{\hb}^{i}(\text{x}_\text{q}), \Tilde{\hb}^{j}(\text{p})),
\end{equation}
where cos-sim denotes cosine similarity.
The predicted target point $\hat{\text{x}}_\text{trg}$ is then determined using the differentiable soft-argmax operation:
\begin{equation}
    \label{eq: predict-target}
    \hat{\text{x}}_\text{trg} = \frac
    {\sum_{\text{p} \in \Omega'} \boldsymbol{S}(\text{p}) \cdot \text{x}_\text{p}}
    {\sum_{\text{p} \in \Omega'} \boldsymbol{S}(\text{p})} \, ,
\end{equation}
where $\Omega' = \{p:  \left\| \text{x}_{\text{p}} - \text{x}_{\text{p}_\text{max}} \right\|_2 \leq R \}$\footnote{The feature maps have a resolution of $44 \times 81$ for an input video resolution of $320 \times 576$, and we set $R=35$.}. 
Thus, the target point prediction can be expressed as
$\hat{\text{x}}_\text{trg} = \mathit{\xi} (\text{x}_\text{q}^i, j, \Tilde{\hb}^{1:N})$, 
and the predicted tracklet for $\text{x}_\text{q}^i$ is given by 
$\mathcal{T}_{\text{x}_\text{q}^i} = \{ \hat{\text{x}}_n : \hat{\text{x}}_n 
= \mathit{\xi} (\text{x}_\text{q}^i, n, \Tilde{\hb}^{1:N}), n=1, ..., N\} $.
Finally, the correspondence loss $\mathcal{L}_{\text{corr}}$ is computed using the Huber loss $L_H$ \citep{huber1992robust}:
\begin{equation}
    \label{eq: corr_loss}
    \mathcal{L}_{\text{corr}}(\Tilde{\hb}^{1:N}, \Omega) = \sum_{(\text{x}_\text{q}^i, \text{x}_\text{trg}^j) \in \Omega} L_H (\mathit{\xi} (\text{x}_\text{q}^i, j, \Tilde{\hb}^{1:N}), \text{x}_\text{trg}^j) 
\end{equation}

When training Track4Gen, we initialize the refiner module as an \textit{identity mapping} to fully leverage the prior of the base model at the start of finetuning. 
To re-route the refined features to the backbone generator, we introduce a trainable \textit{zero convolution} layer \cite{zhang2023adding}, denoted as $\boldsymbol{\zeta}_\psi$.
While the diffusion loss $\mathcal{L}_{\text{diff}}$ back-propagates to all the blocks of the video diffusion model, we detach the gradients of $\Tilde{\hb}^{1:N}$ before passing into $\boldsymbol{\zeta}_\psi$ such that refiner module can solely focus on acquiring the correspondence prior. Hence, given that the output of block $b^k$ is $\hb^{1:N}$, the input to the subsequent block $b^{k+1}$ is computed as  $\hb^{1:N} + \boldsymbol{\zeta}_\psi(\text{stop-gradient}(\boldsymbol{R}_{\phi}(\hb^{1:N})))$.
Fig. \ref{fig: overview} visualizes this architecture design, with red and green colors indicating the objective that optimizes each module.

\begin{figure*}[!t]
    \centering
    \includegraphics[width=\textwidth]
    {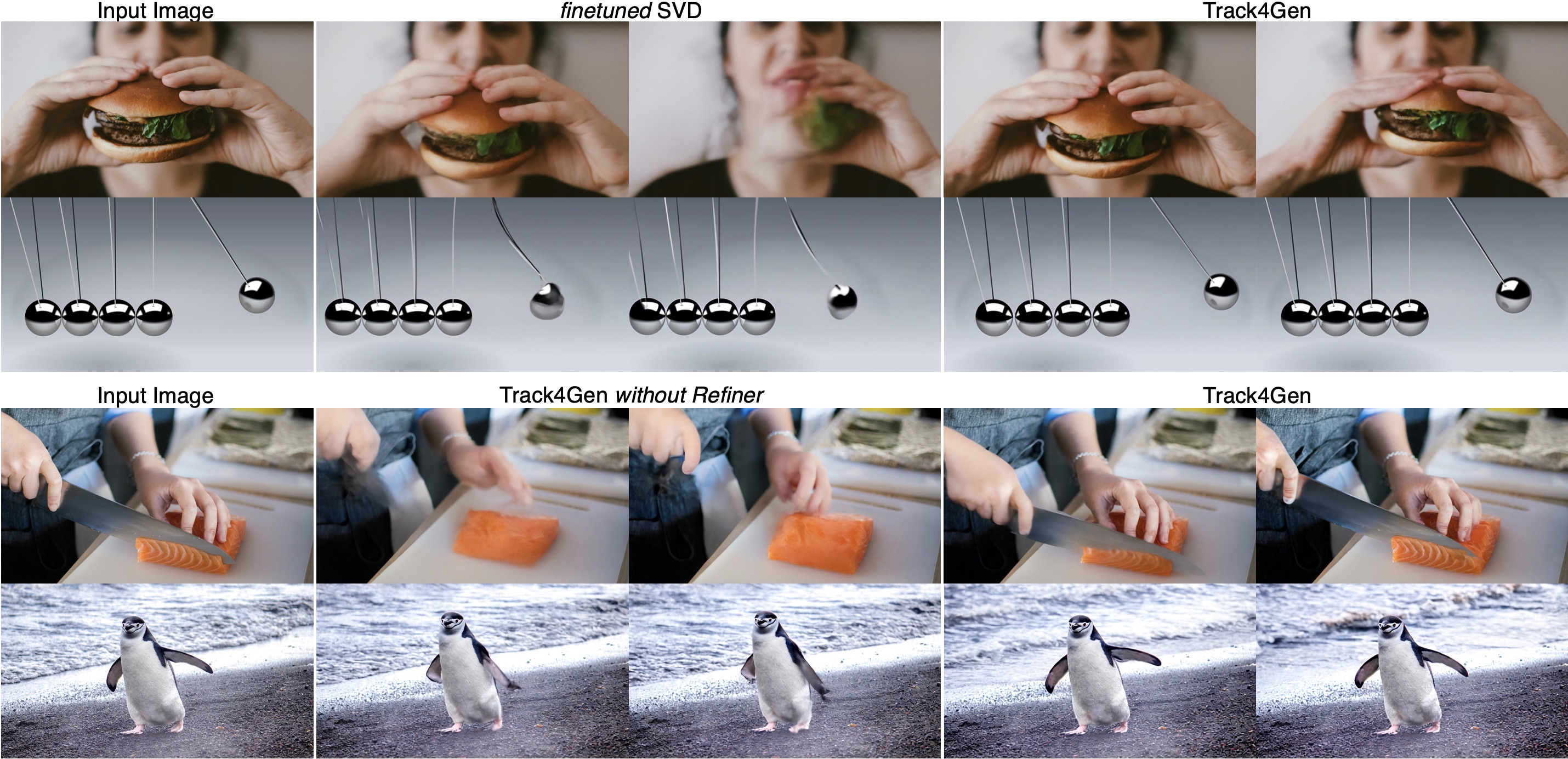}
    \caption{
    \textbf{Qualitative ablation on video generation.}
    Track4Gen is compared with \textit{finetuned} SVD (SVD finetuned on the same training videos without any correspondence supervision) and Track4Gen trained without the Refiner module.
    }
    \label{fig: video-gen-ablation}
\end{figure*}

\section{Experiments}
\subsection{Implementation Details} 
To train Track4Gen, we construct a training dataset consisting of $567$ video-trajectory pairs, with each video having a resolution of $320 \times 576$ and a duration of $24$ frames.
Since no real-world video with (dense) ground-truth trajectory annotations exist at the time of this work, we utilize optical flow to generate trajectory annotations. 
A key challenge is the need for accurate video segmentation maps to ensure a balanced distribution of trajectory points between foreground objects and the background \cite{doersch2022tap}. 
To address this, we utilize public video datasets paired with ground-truth segmentation maps \cite{li2013video, perazzi2016benchmark, pont20172017, caelles20182018, fan2015jumpcut}, where we split longer videos into 24-frame segments.

We use Stable Video Diffusion (SVD) image-to-video pretrained checkpoints\footnote{\url{https://huggingface.co/stabilityai/stable-video-diffusion-img2vid-xt}} as the base video generator.
Our proposed refiner module consists of eight stacked 2D convolution layers and is attached to the third decoder block of the SVD UNet. The refiner module preserves the shape of the hidden states throughout and is initialized as the identity mapping. Further details are provided in the supplementary.
We finetune this enhanced video generator architecture for $20$K steps with our joint loss $\mathcal{L}_{\text{diff}} + \lambda \mathcal{L}_{\text{corr}}$, where $\lambda$ is set to 8.
Rather than finetuning the entire model, we finetune only the temporal transformer blocks, the refiner module $\boldsymbol{R}_{\phi}$, and the zero convolution $\boldsymbol{\zeta}_\psi$.
In each iteration, we sample 512 correspondence pairs from the precomputed trajectories. 
We use the AdamW optimizer \cite{loshchilov2017fixing} with a learning rate of $1e{-}5$, $\beta_1{=}0.9$, $\beta_2{=}0.999$, and a weight decay of $1e{-}2$.
We train the model on $4 {\times}$ H100 GPUs with a total batch size of $4$.
For sampling new videos, we apply the default settings using $30$ steps with the EDM sampler \cite{karras2022elucidating}, \textit{motion bucket id} ${=}127$, and \textit{fps}${=}7$.

\begin{table}[!h]
\centering
\newcommand{\first}[1]{\textbf{#1}}
\newcommand{\second}[1]{\underline{#1}}
\definecolor{Gray}{gray}{0.9}
\caption{
      \textbf{Quantitative comparison on video generation performance}. We compare Track4Gen to the pre-trained SVD $^{\star}$ as well as a finetuned SVD on the same dataset (\textit{finetuned} SVD). We also train a variant of Track4Gen without the refiner module. All videos are generated at 320x576 resolution, except SVD$^{\star}$ (576p) which operates at 576x1024 resolution. 
      }
  \begin{adjustbox}{max width=\columnwidth}
  \begin{tabular}{rccccc|cc}
    \toprule
       & Subject     & Temporal   & Motion     & Imaging & Video-Image  & \multirow{2}{*}{FID} & \multirow{2}{*}{FVD} \\
       & Consistency & Flickering & Smoothness & Quality & Alignment &   &  \\
    \hline
   SVD$^{\star}$      & 0.9535 & 0.9464 & 0.9774 & 0.6648 & 0.9539 & 29.0 & 776 \\
   \textit{finetuned} SVD & 0.9665 & 0.9800 & 0.9909 & 0.6766 & 0.9771 & 27.0 & 735 \\
   Track4Gen \textit{w/o refiner} & 0.9506 & 0.9725 & 0.9791 & 0.6653 & 0.9614 & 27.1 & \first{718} \\
   \rowcolor{Gray}
   Track4Gen & \first{0.9746} & \first{0.9806} & \first{0.9921} & \first{0.6835} & \first{0.9814} & \first{26.6} & 724 \\
   \hdashline
   SVD$^{\star}$ (576p) & 0.9576 & 0.9478 & 0.9795 & 0.6812 & 0.9582 & - & - \\
    \bottomrule
    \end{tabular}
    \end{adjustbox}
  \label{tab: quantitative-video-gen}
\end{table}

\subsection{Track4Gen for Video Generation}
\label{sec: vid-gen-comparison}
We evaluate Track4Gen for the image-to-video generation task via a series of experiments using multiple datasets, automated metrics, and human evaluations.

\textbf{Evaluation Setup.} 
We compare Track4Gen against the original SVD (SVD$^*$) \cite{blattmann2023stable}, as well as a version of SVD that is finetuned on the same videos as Track4Gen (\textit{finetuned} SVD). Furthermore, we train a variant of Track4Gen without the refiner module. 
For VBench metrics \cite{huang2024vbench}, evaluations are conducted on the VBench-I2V dataset, containing 355 diverse images. 
FID and FVD are measured using the DAVIS \cite{pont20172017} dataset as reference.
We generate 24-frame videos conditioned on each input image.

\textbf{Automatic metrics.}
We first report five key metrics from VBench \cite{huang2024vbench}:
(1) \textit{Subject Consistency}—assesses subject appearance consistency of the video by computing the similarity of DINO \cite{oquab2023dinov2} features.
(2) \textit{Temporal Flickering}—detects temporal consistency by taking static frames and calculating the mean absolute difference across frames.
(3) \textit{Motion Smoothness}—measures smoothness of motion, and how well it adheres to real-world physics, using video frame interpolation priors \cite{li2023amt}.
(4) \textit{Image Quality}—evaluates distortions (e.g., noise, blur) using a pretrained, multi-scale image quality predictor \cite{ke2021musiq}.
(5) \textit{Video-Image Alignment}—measures alignment between the subject in the input image and in the generated video using DINO features.
We additionally report FID \cite{heusel2017gans} and FVD \cite{unterthiner2018towards}.

\begin{figure*}[!t]
    \centering
    \includegraphics[width=\textwidth]
    {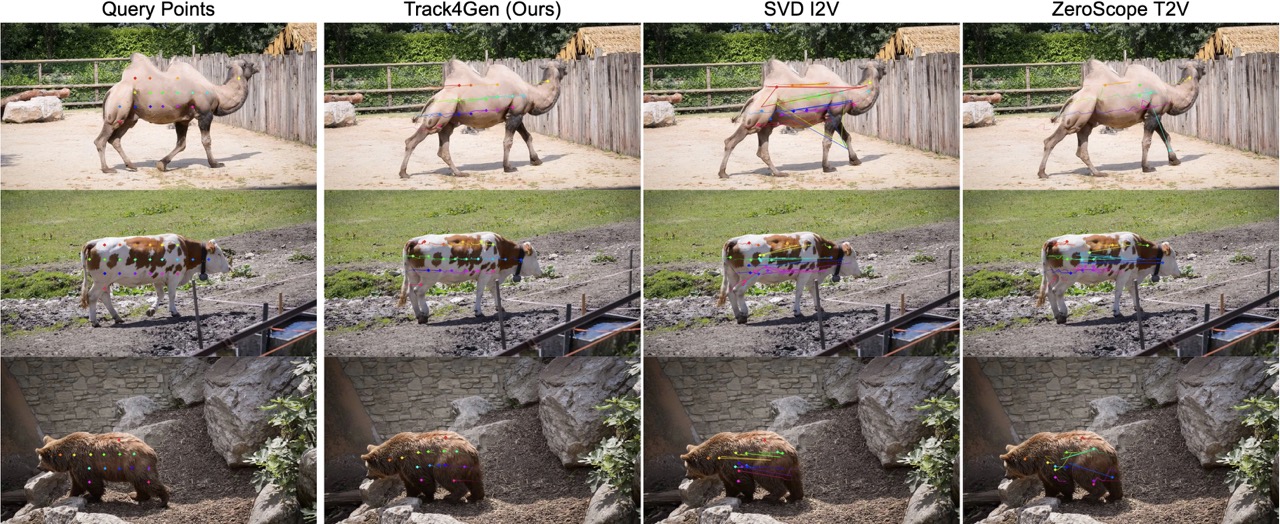}
    \caption{
    \textbf{Qualitative comparison of Track4Gen and baselines for real-world video tracking.} 
    The leftmost column displays query points in the first frame, while the following three columns show tracking results using features from each model.
    }
    \vspace{-2.5mm}
    \label{fig: video-tracking-features-qualitative}
\end{figure*}

\textbf{Human evaluation.}
We further evaluate Track4Gen against baselines through a user study. We ask $64$ participants to compare our results with a randomly selected baseline. We ask the users to evaluate how consistent main objects appear across the frames in a generated video as well as how natural the depicted motion is. We provide further details of the user study in the supplementary material.

\textbf{Qualitative results.}
Qualitative comparisons with the base SVD are shown in Fig. \ref{fig: video-gen-qualitative}. 
As illustrated, Track4Gen generates videos with strong appearance consistency, avoiding issues of appearance drift. 
In contrast, videos produced by the original SVD exhibit noticeable inconsistencies: the sheep's head (row 1) mutates, the plane's wing (row 2) shows unnatural transitions, and the cars (row 3) disappear.
Further comparisons with \textit{finetuned} SVD and Track4Gen without the refiner module are shown in Fig. \ref{fig: video-gen-ablation} and highlight the superior visual coherence of the proposed Track4Gen.

\begin{figure}[t]
     \centering
     \begin{subfigure}[t]{0.47\columnwidth}
         \centering
         \includegraphics[trim={1.3cm 0.0cm 0.9cm 0.0cm},clip,width=\linewidth]{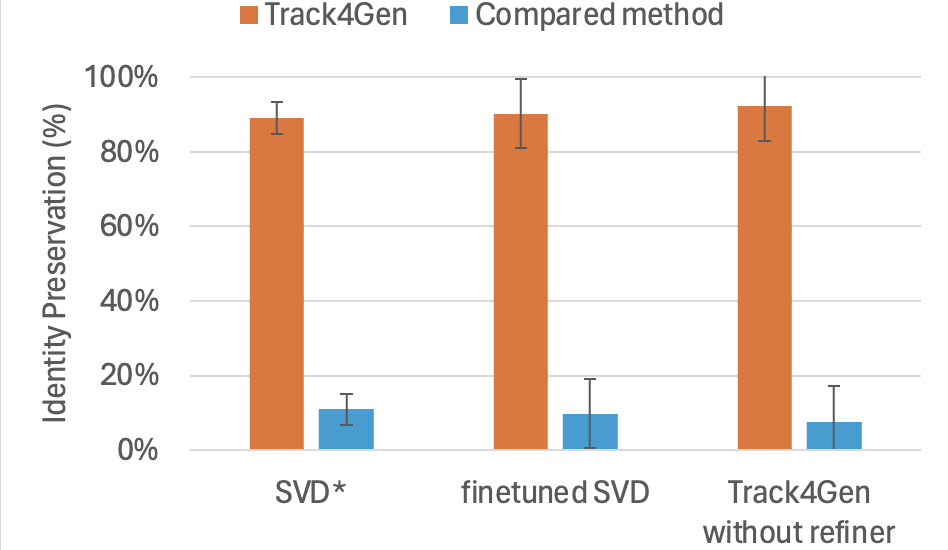}
         \caption{Identity preservation}
         \label{fig:user_identity}
     \end{subfigure}
     \hfill
     \begin{subfigure}[t]{0.47\columnwidth}
         \centering
         \includegraphics[trim={1.2cm 0.0cm 0.9cm 0.0cm},clip,width=\linewidth]{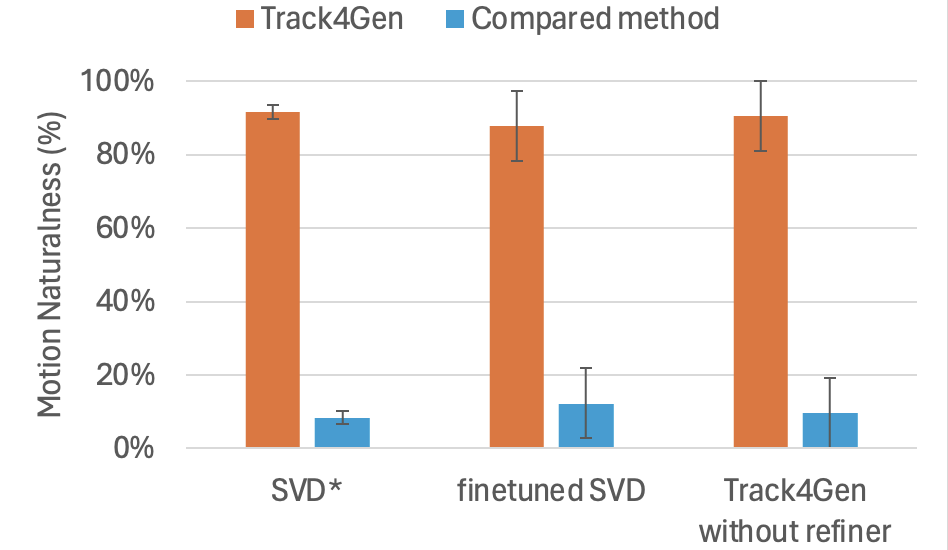}
         \caption{Motion naturalness}
         \label{fig:user_naturalness}
    \end{subfigure}
     \caption{
     \textbf{User study results.}
     Our study shows that Track4Gen better preserves object identity and produces more natural motion.}
     \vspace{-3.3mm}
     \label{fig:user}
\end{figure}

\textbf{Quantitative results.}
As shown in Tab.~\ref{tab: quantitative-video-gen}, our method achieves the highest scores across all 5 metrics from VBench, along with the lowest FID and second-lowest FVD values, outperforming the base SVD by substantial margins. Fig.~\ref{fig:user} provides the user study results where the majority of the participants agreed that Track4Gen is superior both in terms of identity preservation and naturalness of motion.

\subsection{Track4Gen for Video Tracking}
\vspace{-0.5mm}
We evaluate Track4Gen's capability to \textit{track any point} in real videos by adding a small amount of noise to the input video \cite{tang2023emergent} and passing it through the video denoiser $\fb_{\theta}$ to extract feature maps. We first compare tracking results with such features against other raw features \cite{teed2020raft, zeroscope, blattmann2023stable} in Sec. \ref{sec: raw-features-comparisons}. In Sec. \ref{sec: with-test-time-optimization}, we utilize Track4Gen's features in a test-time optimization method \cite{tumanyan2024dino} and compare to both self-supervised and fully supervised video trackers.

\begin{table}[!t]
\centering
\newcommand{\first}[1]{\textbf{#1}}
\newcommand{\second}[1]{\underline{#1}}
\definecolor{Gray}{gray}{0.9}
\caption{
      \textbf{Quantitative zero-shot feature comparison on video tracking benchmarks}.
      Track4Gen features are compared to the features of SVD$^{*}$ \cite{blattmann2023stable},  ZeroScope \cite{zeroscope}, and RAFT \cite{teed2020raft}. For all the metrics, higher values indicate better performance.
      }
  \begin{adjustbox}{max width=\columnwidth}
  \begin{tabular}{r|c|c|c|c|c||c|c|c|c|c}
    \toprule
       & \multicolumn{3}{c|}{DAVIS-480p} & \multicolumn{2}{c||}{BADJA} & \multicolumn{3}{c|}{DAVIS-480p} & \multicolumn{2}{c}{BADJA}\\
       \multirow{3}{*}{Method} & \multicolumn{3}{c|}{(24-frame)} & \multicolumn{2}{c||}{(24-frame)} & \multicolumn{3}{c|}{(whole duration)} & \multicolumn{2}{c}{(whole duration)}\\
      & $\delta^{x}_{avg}$ & OA & AJ & $\delta^{\textit{seg}}$ & $\delta^{3px}$ & $\delta^{x}_{avg}$ & OA & AJ & $\delta^{\textit{seg}}$ & $\delta^{3px}$\\
    \hline
    ZeroScope & 46.2 & 67.0 & 39.4 & 27.5 & 2.8 & 37.2 & 59.5 & 27.8 & 19.9 & 2.0 \\
    SVD$^{*}$ & 42.4 & 79.7 & 36.4 & 26.2 & 2.9 & 35.4 & 70.1 & 26.5 & 19.4 & 2.2 \\
    \rowcolor{Gray}
    Track4Gen & \second{69.7} & \first{85.8} & \first{56.5} & \second{52.3} & \second{7.7} & \second{58.9} & \first{78.4} & \first{40.2} & \second{40.4} & \second{5.0} \\
    \hdashline
    RAFT & \first{73.3} & - & - & \first{54.8} & \first{8.7} & \first{66.7} & - & - & \first{45.0} & \first{5.8} \\
    \bottomrule
    \end{tabular}
    \end{adjustbox}
  \label{tab: video-tracking-quantiative-raw-features}
  \vspace{-3.3mm}
\end{table}

\vspace{-0.5mm}
\subsubsection{Zero-shot Feature Comparison}
\label{sec: raw-features-comparisons}

We evaluate the precision of predicted tracks using the features from Track4Gen, the original SVD model (SVD$^{\star}$), and RAFT \cite{teed2020raft}. We also test another text-to-video model, ZeroScope T2V \cite{zeroscope}, to demonstrate how raw features from pre-trained video generators typically work out of the box.
For RAFT, tracking is achieved by chaining optical flow displacements, while the others use nearest neighbor matching between its encoded features.

\textbf{Datasets.}
We use TAP-Vid DAVIS~\cite{doersch2022tap} and BADJA~\cite{biggs2019creatures} as benchmark datasets.
Additionally, we include two shorter benchmarks, DAVIS (24-frame) and BADJA (24-frame), which focus on the first 24 frames with query and target points within this range. Details on encoding long videos with the video models are in the supplemental.

\textbf{Metrics.}
For evaluating the TAP-Vid benchmarks, we use the following metrics: (i) Position Accuracy ($\delta^{x}_{avg}$) evaluates the average accuracy of visible points, where each $\delta^{x}$ represents the fraction of predicted points that lie within $x$ pixels of the ground-truth position, with $x \in \{1, 2, 4, 8, 16\}$. (ii) Occlusion Accuracy (OA)  evaluates the correctness of occlusion predictions. (iii) Average Jaccard (AJ) jointly assesses both position and occlusion accuracy.
For the BADJA dataset, we report $\delta^{\textit{seg}}$, which measures the accuracy of tracked keypoints within a distance of $0.2\sqrt{A}$ from the ground-truth annotation, where $A$ is the area of the foreground object. We also report $\delta^{3px}$, which assesses accuracy within a 3-pixel threshold. 
A cosine similarity threshold of $0.6$ is used for occlusion prediction.

\textbf{Results.}
We present the qualitative results in Fig. \ref{fig: video-tracking-features-qualitative} and the quantitative results in Tab. \ref{tab: video-tracking-quantiative-raw-features}. Although primarily designed for video generation, Track4Gen boosts the poor performance of the pre-trained video models significantly, approaching the accuracy of RAFT optical flow chaining.

\begin{figure}[!t]
    \centering
    \includegraphics[width=\columnwidth]
    {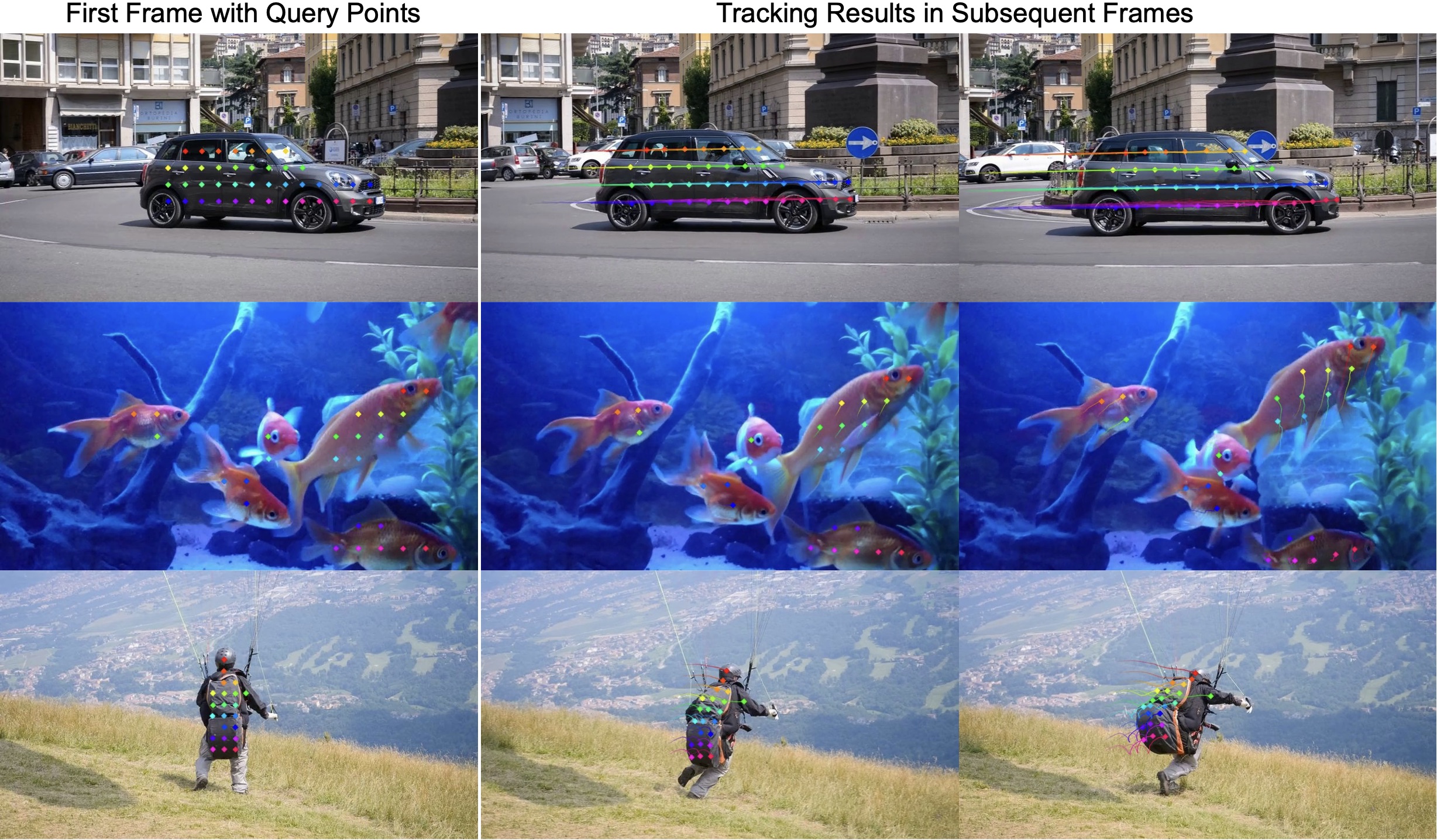}
    \caption{
    \textbf{Extending Track4Gen with test-time adaptation \cite{tumanyan2024dino}}.
    }
    \vspace{-1mm}
    \label{fig: ours-with-dino-tracker}
\end{figure}

\subsubsection{Extending Track4Gen with Test-time Adaptation}
\label{sec: with-test-time-optimization}

To further evaluate Track4Gen's long-term tracking capabilities, we integrate our features with test-time adaptation algorithm of DINO-Tracker \cite{tumanyan2024dino}, where a per-video optimization is performed using optical flow supervision. We replace the originally used DINOv2 \cite{oquab2023dinov2} with the features from Track4Gen.
We evaluate using the same datasets and metrics outlined in Sec. \ref{sec: raw-features-comparisons}, against both fully-supervised trackers \cite{doersch2022tap, zheng2023pointodyssey, doersch2023tapir} and self-supervised methods \cite{wang2023tracking, tumanyan2024dino}.

Tab. \ref{tab: video-tracking-quantiative-test-time} shows that Track4Gen features optimized with \cite{tumanyan2024dino} achieve performance comparable to dedicated trackers.
Qualitative results are in Fig. \ref{fig: ours-with-dino-tracker} 
and in the supplemental.

\begin{table}[t!]
\newcommand{\first}[1]{\textbf{#1}}
\newcommand{\second}[1]{\underline{#1}}
\definecolor{Gray}{gray}{0.9}
\caption{
      \textbf{Quantitative comparison with video trackers}.
      Although primarily designed for \textit{video generation}, Track4Gen combined with a test-time optimization method \cite{tumanyan2024dino} achieves performance comparable to dedicated \textit{video tracking} frameworks, even when compared to supervised methods. 
      }
  \begin{adjustbox}{max width=\columnwidth}
  \begin{tabular}{r|c|c|c|c|c}
    \toprule
     \multirow{2}{*}{Method} & \multicolumn{3}{c|}{DAVIS-480} & \multicolumn{2}{c}{ BADJA} \\
     & $\delta^{x}_{avg}$ & OA & AJ & $\delta^{\textit{seg}}$ & $\delta^{3px}$ \\
    \hline
    TAP-Net$^{\star}$ & 66.4 & 79.0 & 46.0 &  45.4 & 9.6 \\
    PIPs++$^{\star}$ & 73.6 & - & - &  59.0 & 9.8 \\
    TAPIR$^{\star}$ & \second{77.3} & \first{89.5} & \first{65.7} & \second{68.7} & 10.5 \\
    \hline
    Omnimotion$^{\dagger}$ & 74.1 & 84.5 & 58.4 & 45.2 & 6.9 \\    
    DINO-Tracker$^{\dagger}$ & \first{80.4} & \second{88.1} & \second{64.6} & \first{72.4} & \first{14.3} \\
    \rowcolor{Gray}
    DINO-Tracker w/ Track4Gen$^{\dagger}$ & 72.5 & 84.5 & 55.7 & 48.4 & \second{10.9} \\
    \bottomrule
    \end{tabular}
    \end{adjustbox}
{\scriptsize $^{\star}$ -- supervised. \quad $^{\dagger}$ -- test-time training.}
\vspace{-2mm}
  \label{tab: video-tracking-quantiative-test-time}
\end{table}

\begin{table}[!t]
\centering
\newcommand{\first}[1]{\textbf{#1}}
\newcommand{\second}[1]{\underline{#1}}
\definecolor{Gray}{gray}{0.9}
\caption{
      \textbf{Ablation on trainable modules and refiner}. 
      }
  \begin{adjustbox}{max width=\columnwidth}
  \begin{tabular}{rccccc}
    \toprule
     Trainable  & Subject    & Temporal & Motion     & Imaging & Video-Image \\
     modules     & Consistency & Flickering & Smoothness & Quality & Alignment   \\
    \hline
   spatial ${+}$ temporal & 0.9734 & \first{0.9811} & 0.9917 & \first{0.6863} & 0.9807 \\
   spatial & 0.9726 & 0.9801 & 0.9919 & 0.6852 & 0.9811 \\
   temporal & \first{0.9746} & 0.9806 & \first{0.9921} & 0.6835 & \first{0.9814} \\
   \toprule
   Refiner architecture &  &  &  &  &  \\
   \hline
   2D convolutions & \first{0.9746} & \first{0.9806} & \first{0.9921} & \first{0.6835} & 0.9814 \\
   3D convolutions & 0.9687 & 0.9734 & 0.9904 & 0.6833 & \first{0.9820}\\
    \bottomrule
    \end{tabular}
    \end{adjustbox}
    \vspace{-1mm}
  \label{tab: quantitative-ablation}
\end{table}

\begin{table}[!t]
\centering
\newcommand{\first}[1]{\textbf{#1}}
\newcommand{\second}[1]{\underline{#1}}
\definecolor{Gray}{gray}{0.9}
\caption{
      \textbf{Quantitative ablation on using annotated, but synthetic videos \cite{greff2022kubric}}. 
      \textit{Left}: Video generation metrics.
      \textit{Right}: Video tracking metrics.
      }
  \begin{adjustbox}{max width=\columnwidth}
  \begin{tabular}{rccc|cc}
    \toprule
      Dataset         & Subject      & Motion   & Imaging   & \multicolumn{2}{c}{BADJA} \\
      composition     & Consistency  & Smoothness & Quality & $\delta^{\textit{seg}}$ & $\delta^{3px}$  \\
    \hline
   real videos & \first{0.9747} & \first{0.9921} & \first{0.6833} & 40.4 & \first{5.0} \\
   real ${+}$ synthetic videos  & 0.9708 & 0.9892 & 0.6793 & \first{42.1} & 4.8 \\
    \bottomrule
    \end{tabular}
    \end{adjustbox}
    \vspace{-2mm}
  \label{tab: quantitative-ablation-dataset}
\end{table}

\subsection{Ablation Studies}
\label{sec:ablation}

We present an ablation study in Tab. \ref{tab: quantitative-ablation} where we train different set of modules.
Each spatio-temporal block of SVD includes both spatial and temporal transformers.
We compare training only spatial transformers, only temporal transformers, or both.
We also ablate the architecture of the refiner module using either 2D or 3D convolution layers.
Our analysis shows that while results are similar across settings, training only the temporal transformers in SVD with 2D convolutions as the refiner module yields optimal video generation quality.
We further analyze our training dataset by additionally incorporating Kubric \cite{greff2022kubric} simulated videos (1K video-track pairs from the Panning MOVi-E data \cite{doersch2023tapir,cho2024local}) with automatically annotated trajectories into training.
As shown in Tab. \ref{tab: quantitative-ablation-dataset}, optical flow-chained tracklets from real provides provide as effective correspondence guidance as tracklets from synthetic data, while synthetic videos negatively impact the video generation quality.
\section{Conclusion and Future Work}
We have presented the first unified framework that bridges two distinct tasks:  video generation and dense point tracking.
We demonstrated that this produces temporally consistent feature representations and appearance-consistent videos.
As for limitations, videos generated by Track4Gen tend to exhibit less dynamic motion compared to those from other video generators. 
Additionally, failure cases are included in the supplementary material.

\textbf{Future work.} 
Recently, cutting-edge video trackers \cite{cho2024local, doersch2024bootstap, karaev2024cotracker3} have emerged, enabling dense, accurate, and long-term tracking, especially with better handling of occlusions.
This opens up promising future directions for extending our work to utilize real-world videos, automatically annotated by these trackers to produce 3D-consistent videos~\cite{huang2025unifyingvideogenerationcamera}.

\textbf{Acknowledgments.}
We thank Seokju Cho and Narek Tumanyan for their invaluable feedback on video point tracking.
We also extend our gratitude to Mingi Kwon, Joon-Young Lee, and Gabriel Huang for their insightful discussions.
Hyeonho Jeong and Jong Chul Ye are supported by the National Research Foundation of Korea (NRF) under Grants RS-2024-00336454 and RS-2023-00262527, and by the Institute for Information \& Communications Technology Planning \& Evaluation (IITP) grant funded by the Korea government (MSIT) (RS-2019-II190075, Artificial Intelligence Graduate School Program, KAIST).
Niloy J. Mitra is supported by UCL AI Centre.


{
    \small
    \bibliographystyle{ieeenat_fullname}
    \bibliography{main}
}

\clearpage
\begin{appendix}
\maketitlesupplementary

\definecolor{calpolypomonagreen}{rgb}{0.12, 0.3, 0.17}

This supplementary material is structured as follows:
Sec. \ref{supple-sec: experimental details} provides additional implementation details for the experiments.
In Sec. \ref{supple-sec: additional metrics}, we report supplementary quantitative metrics for video generation assessment.
Sec. \ref{supple-sec: additional video generation} presents additional qualitative results for image-to-video generation, while Sec. \ref{supple-sec: additional video tracking} focuses on qualitative video tracking results.
Following this, we discuss the potential limitations and failure cases of Track4Gen in Sec. \ref{supple-sec: limitation}.

A comprehensive view of results in the form of videos is available on our \href{https://hyeonho99.github.io/track4gen}{project page}.
Furthermore, an extensive video generation comparison against all baselines can be found on \href{https://hyeonho99.github.io/track4gen/full.html}{this page}.

\section{Experimental Details}
\label{supple-sec: experimental details}

\subsection{Preprocessing Video Correspondence}
We utilize RAFT optical flow \cite{teed2020raft} to compute dense point trajectories across video frames. 
RAFT has demonstrated robust point tracking performance across various input types \cite{wang2023tracking}, even compared to supervised trackers like TAP-Net \cite{doersch2022tap}.
Following previous tracking literature \cite{wang2023tracking, tumanyan2024dino},
we first compute pairwise correspondences between all consecutive frames.
Tracks are then formed by chaining the estimated flow fields and filtered using a cycle consistency constraint.
Specifically, given a point $\boldsymbol{\text{x}}^i$ in frame $i$ and optical flow between frames $i$ and $i+1$ denoted as $\boldsymbol{f}_{i \rightarrow i+1}$, the corresponding point in frame $i+1$ is estimated as $\boldsymbol{\text{x}}^{i+1} = \boldsymbol{\text{x}}^i + \boldsymbol{f}_{i \rightarrow i+1}(\boldsymbol{\text{x}}^i)$.
We retain the pair $(\boldsymbol{\text{x}}^i, \boldsymbol{\text{x}}^{i+1})$ only if it satisfies $\| \boldsymbol{\text{x}}^i - (\boldsymbol{\text{x}}^{i+1} + \boldsymbol{f}_{i+1 \rightarrow i}(\boldsymbol{\text{x}}^{i+1})) \|_2 \le 1.5$, where $h{\times}w$ is set as $320{\times}576$.
Also, a pair $(\boldsymbol{\text{x}}^i, \boldsymbol{\text{x}}^{j})$ is filtered out
if $\| \boldsymbol{\text{x}}^j {-} \boldsymbol{\text{x}}^{i \rightarrow j} \|_2 {\ge 2}$ 
and
$\| \boldsymbol{\text{x}}^i {-} (\boldsymbol{\text{x}}^{i \rightarrow j} {+} \boldsymbol{f}_{j \rightarrow i}(\boldsymbol{\text{x}}^{i \rightarrow j})) \|_2 {\le 1.5}$.

\subsection{Refiner Network}
When training Track4Gen, we design a convolutional neural network for the refiner module $\boldsymbol{R}_{\phi}$.
The network comprises 8 layers, each with a fixed channel dimension of 640, a kernel size of 3, stride of 1, and padding of 1.
The first 7 layers follow the structure \texttt{Conv2d $\rightarrow$ BatchNorm2d $\rightarrow$ ReLU}, except for the last layer which consists of \texttt{Conv2d $\rightarrow$ ReLU}.

\begin{figure}[!t]
    \centering
    \includegraphics[width=\columnwidth]{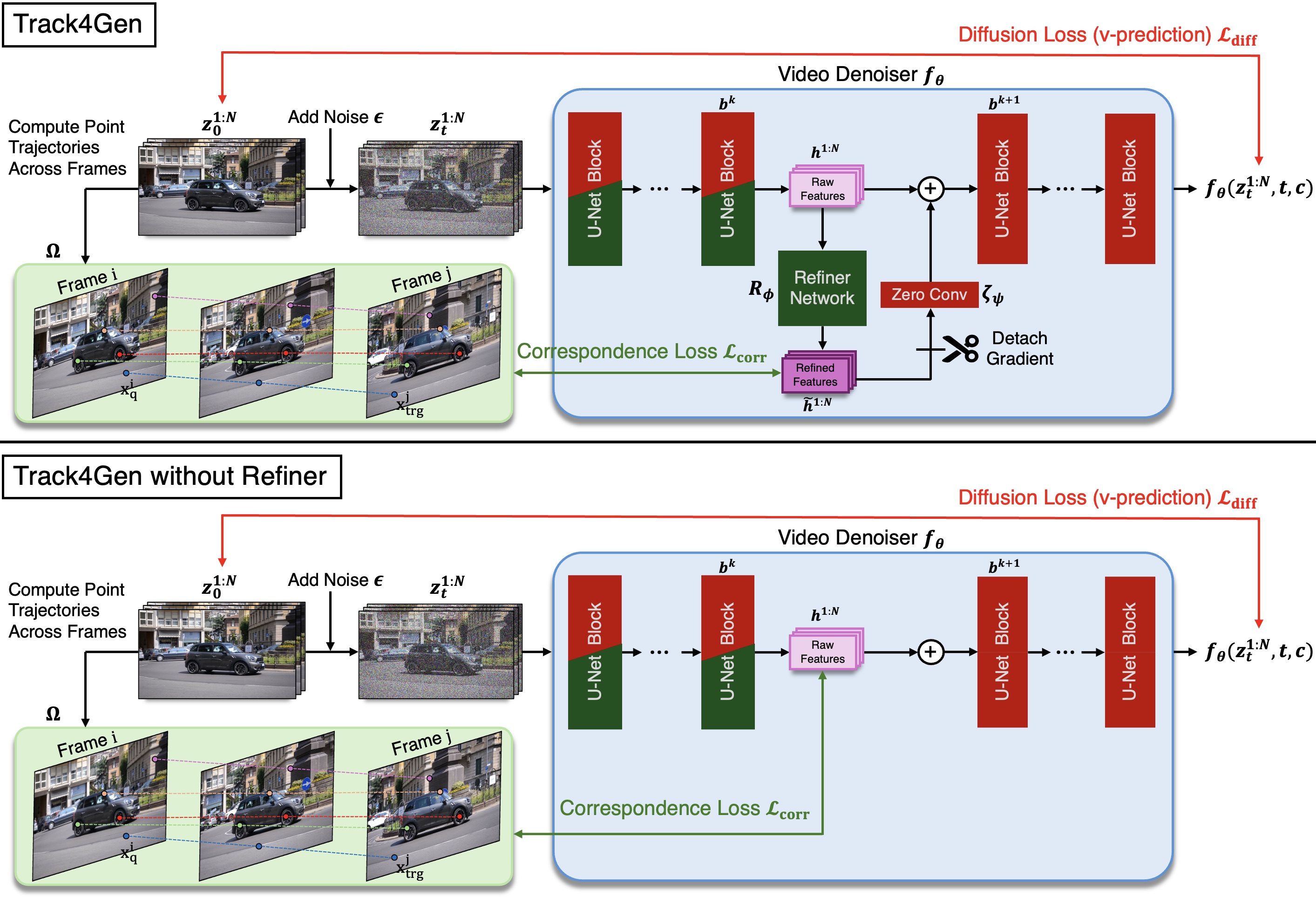}
    \caption{
    \textbf{Comparison of Track4Gen with and without Refiner.}
    \textit{Top:} Correspondence loss $\mathcal{L}_{\text{corr}}$ is computed using the refined features $\Tilde{\hb}^{1:N}$.
    \textit{Bottom:} Correspondence loss $\mathcal{L}_{\text{corr}}$ is computed using the raw diffusion features $\hb^{1:N}$.
    }
    \label{fig: without-refiner}
\end{figure}

To better demonstrate the architecture of the baseline \textit{Track4Gen without Refiner}, we provide a visualization in Fig. \ref{fig: without-refiner}. The figure compares the training schemes of this baseline with Track4Gen.
In this variant, the correspondence loss $\mathcal{L}_{\text{corr}}$ is computed directly from the raw video diffusion features $\hb^{1:N}$.

\subsection{User Study Details}
\begin{figure}[!t]
    \centering
    \includegraphics[width=\columnwidth]{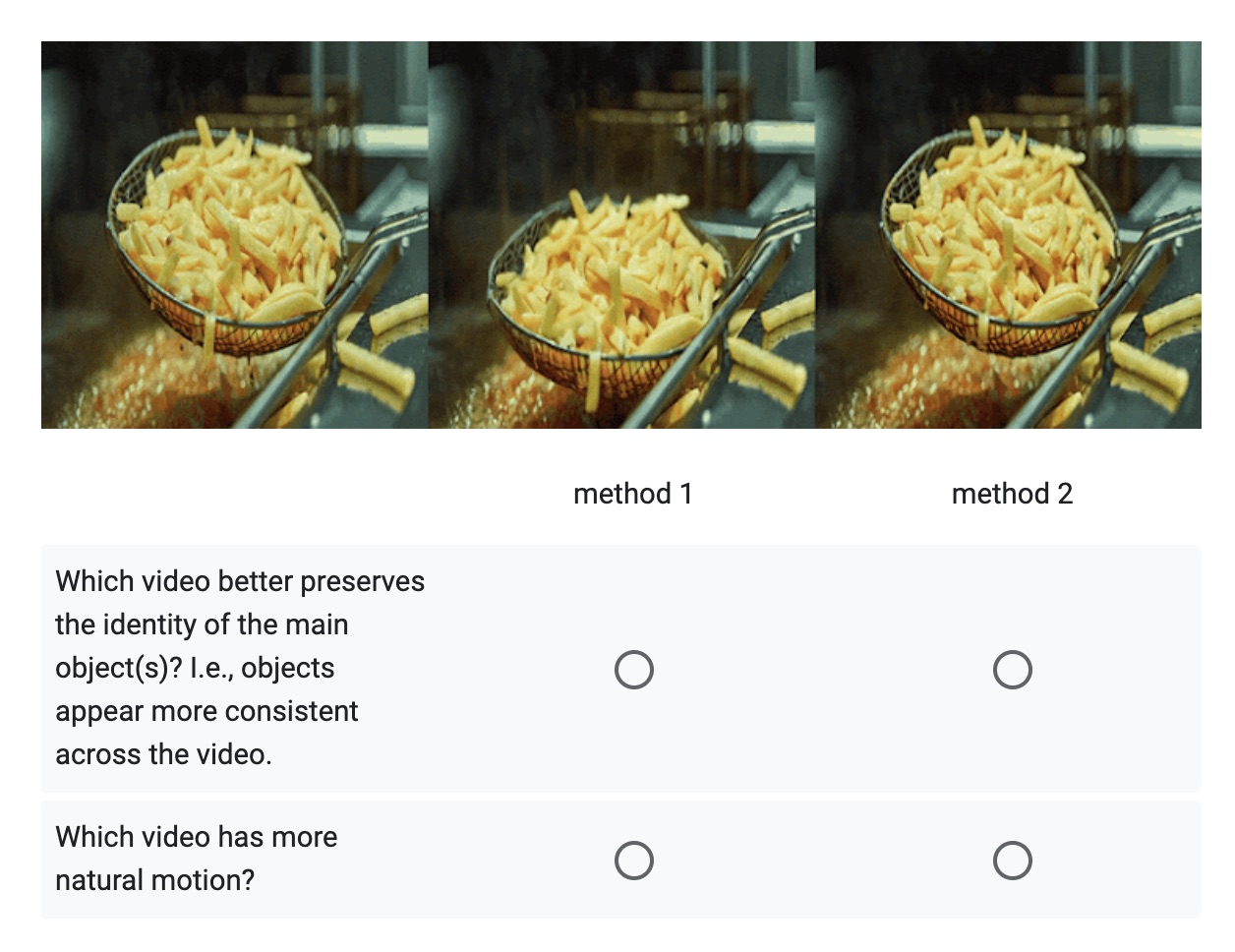}
    \caption{
    \textbf{Example user evaluation page.} 
    The order of Track4Gen and the baseline is randomly shuffled to ensure a fair comparison.
    }
    \label{fig: supple-user-study}
\end{figure}

Fig.~\ref{fig: supple-user-study} shows an example of our user evaluation page. 
The input image is displayed on the left, while the middle and right columns show two generated videos for comparison. 
One result is from Track4Gen, and the other is randomly selected from four baselines: pretrained Stable Video Diffusion \cite{blattmann2023stable}, finetuned Stable Video Diffusion without correspondence supervision, and Track4Gen trained without the refiner module.
Note that the order of Track4Gen and the baseline is randomly shuffled (i.e., Track4Gen may appear first or the baseline may appear first).
Participants are asked to answer two questions:
(i)~\textit{Identity preservation}: Which video better preserves the identity of the main object(s)?
(ii)~\textit{Motion naturalness}: Which video has more natural motion?

\subsection{Encoding Long Videos with Video Diffusion Models}
Majority of video diffusion models struggle with flexibility in temporal resolution.
Specifically, if a model is trained on a fixed temporal resolution of $N$ frames (e.g., $N=24$), the quality of generated videos significantly degrades when attempting to generate videos with a much larger number of frames. 
Similarly, when these models are used as video feature extractors, the extracted features are invalid if the input video contains significantly more frames than the model was trained to handle.

This limitation poses a challenge, as most videos in video tracking benchmarks contain more frames than the training resolution of video diffusion models.
To address this, for a benchmark video with temporal resolution $M$, where $M \gg N$, we split the $M$-frame video into $N$-frame segments and encode each segment independently.
For the final segment, which may contain fewer than $N$ frames, we extend it by borrowing frames from the previous segment. For instance, if the last segment is 14 frames long and $N=24$, we append the last 10 frames from the previous segment to complete the sequence.
This extended segment is then passed through the video diffusion model to extract features. After encoding, we discard the features of the the borrowed frames, retaining only the features for the original frames in the segment.

\begin{table}[!t]
\centering
\newcommand{\first}[1]{\textbf{#1}}
\newcommand{\second}[1]{\underline{#1}}
\definecolor{Gray}{gray}{0.9}
\caption{
      \textbf{CLIP similarity and LPIPS comparison for assessing temporal consistency}.
      We compare Track4Gen to the pre-trained SVD as well as a finetuned SVD on the same dataset (\textit{finetuned} SVD), and a variant of Track4Gen without the refiner module.
      }
  \begin{adjustbox}{max width=\columnwidth}
  \begin{tabular}{rcc}
    \toprule
       & CLIPSIM $\uparrow$ & LPIPS $\downarrow$ \\
    \hline
   Pretrained SVD      & 0.9839 & 0.1373 \\
   \textit{finetuned} SVD & 0.9869 & 0.0913 \\
   Track4Gen \textit{without refiner} & 0.9923 & 0.0547 \\
   \rowcolor{Gray}
   Track4Gen & \first{0.9924} & \first{0.0533} \\
    \bottomrule
    \end{tabular}
    \end{adjustbox}
  \label{tab: supple-clip-lpips}
\end{table}

\section{Additional Metrics}
\label{supple-sec: additional metrics}
To further evaluate the temporal consistency of generated videos, we report CLIPSIM \cite{radford2021learning} and LPIPS \cite{zhang2018unreasonable} metrics.
For CLIPSIM, we compute the average CLIP similarity between all neighboring frame pairs using the CLIP Image Encoder.
Similarly, we calculate the average LPIPS distance between neighboring frame pairs to assess perceptual differences.
As shown in Tab. \ref{tab: supple-clip-lpips}, Track4Gen achieves the highest CLIP similarity and lowest LPIPS distance, demonstrating its superior temporal consistency in the videos it generates.

\section{Additional Video Generation Results}
\label{supple-sec: additional video generation}

\subsection{Comparisons}
\begin{figure*}[!htb]
    \centering
    \includegraphics[width=\textwidth]{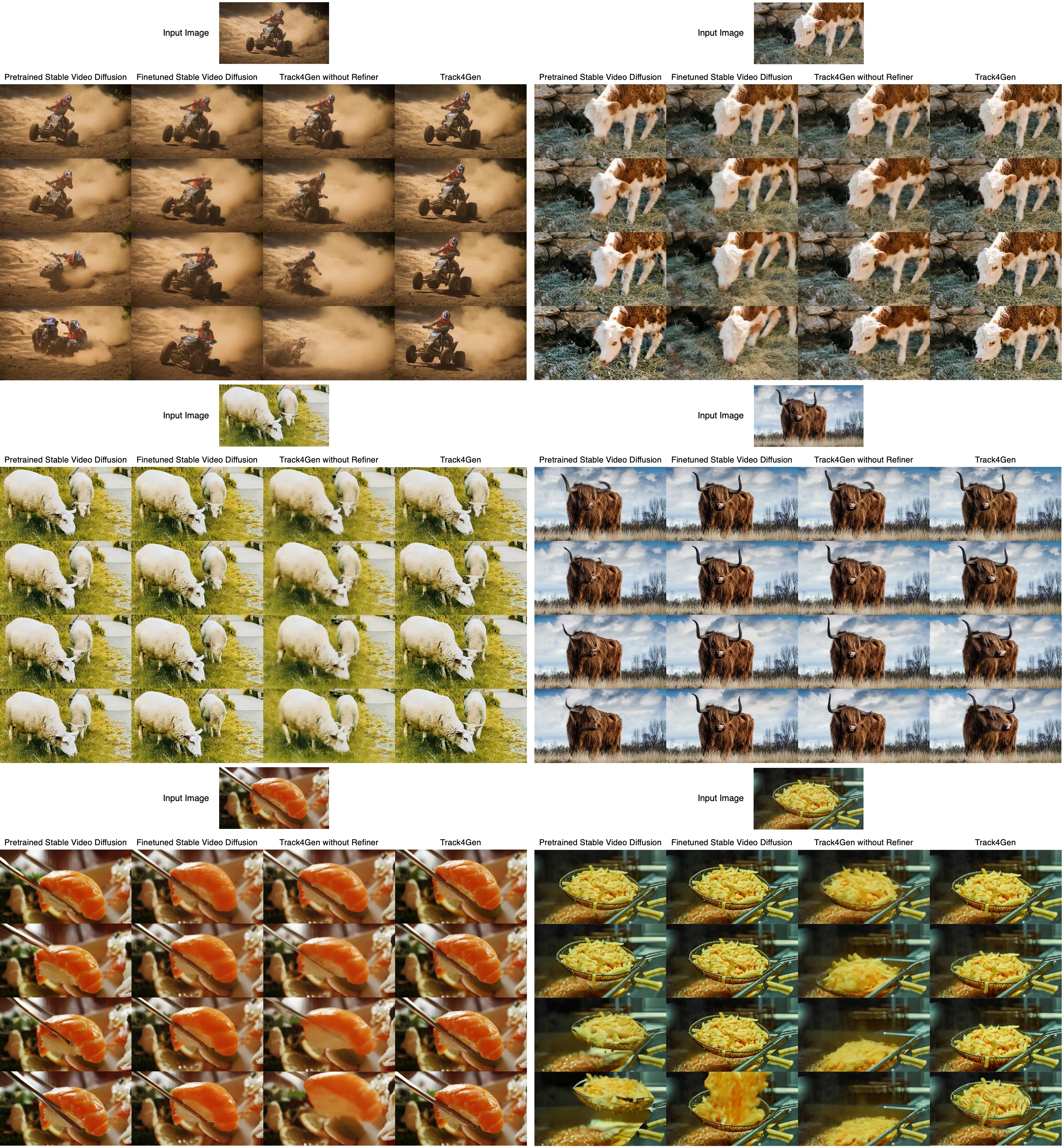}
    \caption{
    \textbf{Qualitative video generation results:}
    Track4Gen compared against all three baselines.
    }
    \label{fig: supple-vid-gen-4-1}
\end{figure*}

\begin{figure*}[!htb]
    \centering
    \includegraphics[width=\textwidth]{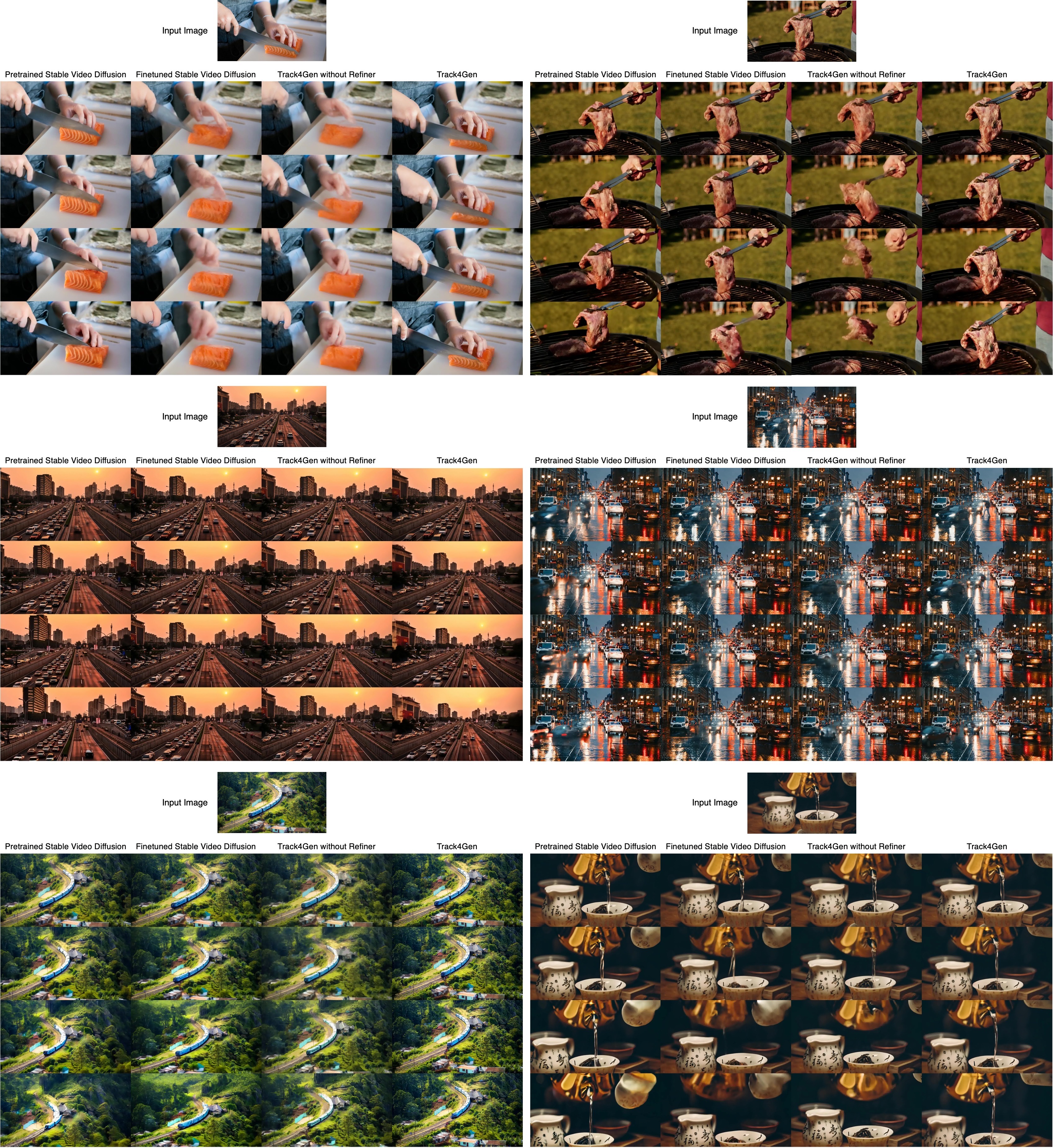}
    \caption{
    \textbf{Qualitative video generation results:}
    Track4Gen compared against all three baselines.
    }
    \label{fig: supple-vid-gen-4-2}
\end{figure*}

In Fig. \ref{fig: supple-vid-gen-4-1} and \ref{fig: supple-vid-gen-4-2}, we present a comparison of Track4Gen against all three baselines: (1) the pretrained Stable Video Diffusion, (2) Stable Video Diffusion finetuned without the tracking loss, and (3) Track4Gen trained without the Refiner module.
For a better view, please visit \textit{page 2 (top)} of our project page.

\subsection{Video Generation with Embedded Tracks}
\begin{figure*}[!htb]
    \centering
    \includegraphics[width=\textwidth]{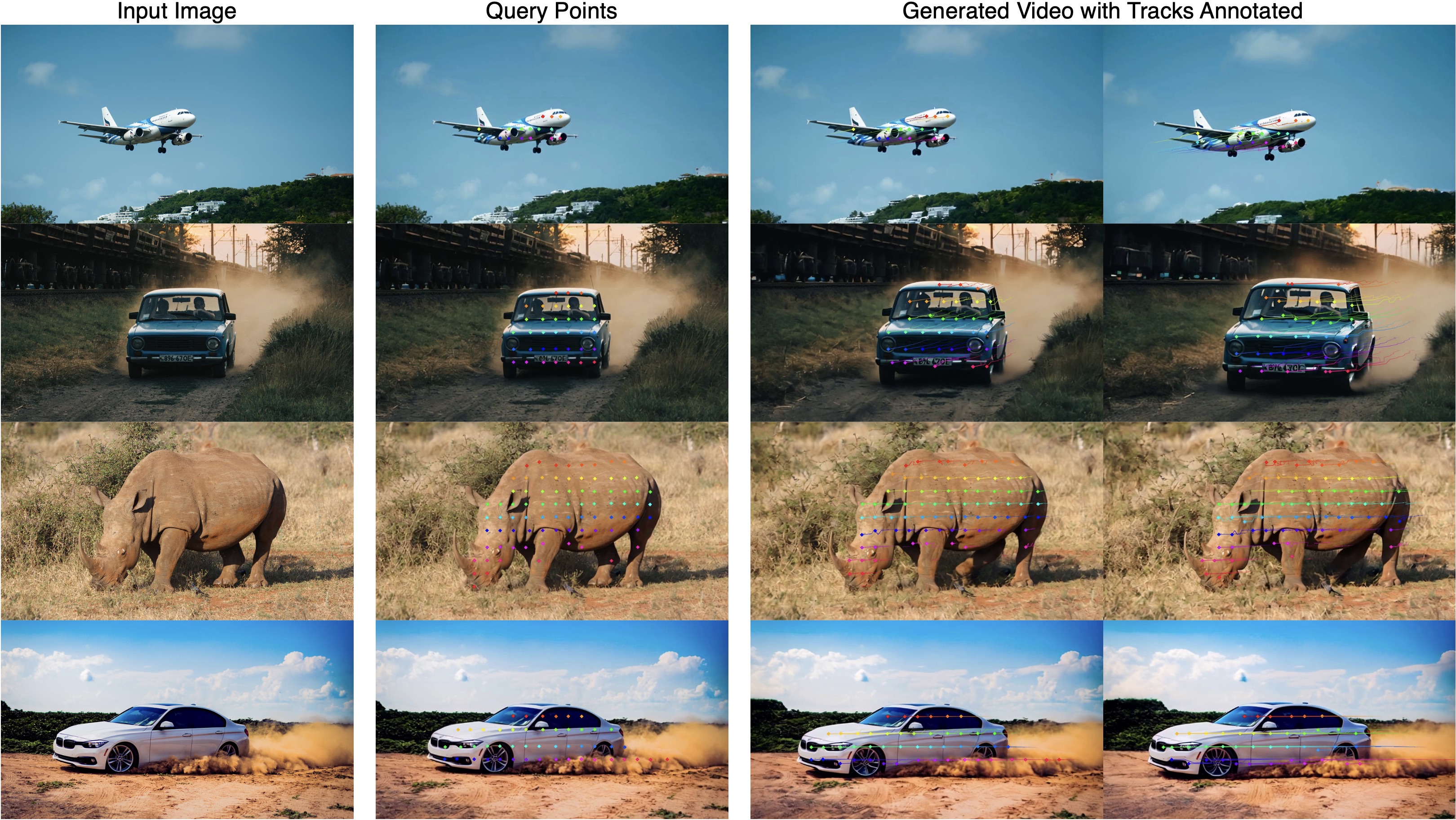}
    \caption{
    \textbf{Generated Videos with Embedded Tracks.}
    Predicted point tracks are annotated on the videos generated by Track4Gen.
    }
    \label{fig: supple-generated-tracking}
\end{figure*}

To demonstrate that Track4Gen generates videos with temporally consistent feature representations, we visualize the predicted point tracks annotated on the generated videos in Fig. \ref{fig: supple-generated-tracking}. These tracks are computed in a zero-shot setting, using the intermediate features extracted from the final denoising step.

\section{Additional Video Tracking Results}
\label{supple-sec: additional video tracking}

\subsection{Feature Comparisons}
\begin{figure*}[!htb]
    \centering
    \includegraphics[width=\textwidth]{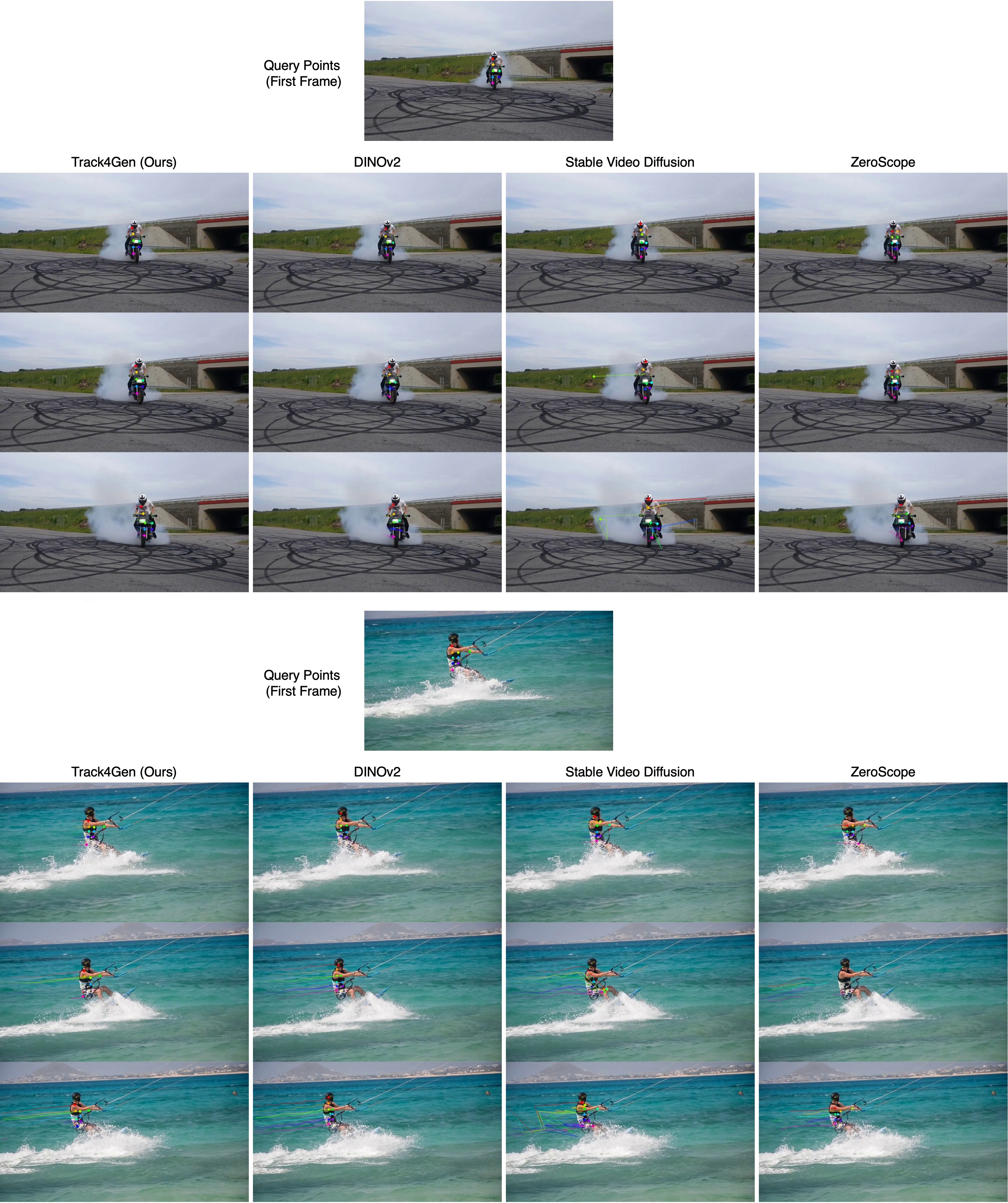}
    \caption{
    \textbf{Additional feature comparison on real-world video tracking:}
    Track4Gen vs DINOv2 vs Stable Video Diffusion vs ZeroScope
    }
    \label{fig: supple-feature-track}
\end{figure*}

\begin{figure*}[!htb]
    \centering
    \includegraphics[width=\textwidth]{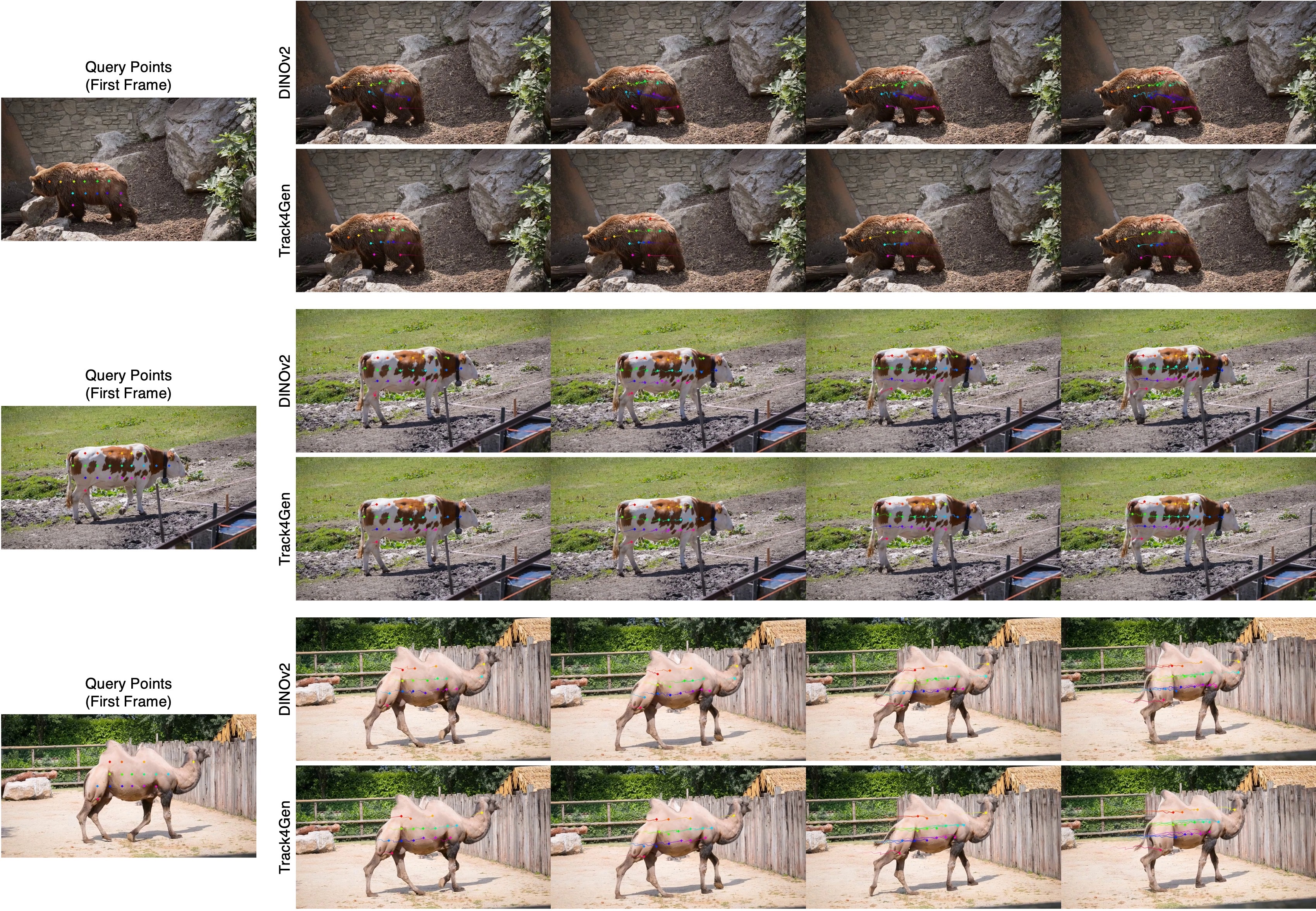}
    \caption{
    \textbf{Additional feature comparison on real-world video tracking:}
    Track4Gen vs DINOv2
    }
    \label{fig: supple-vs-dino}
\end{figure*}

DINO features \cite{caron2021emerging, oquab2023dinov2} are widely recognized for their accuracy in image correspondence tasks \cite{amir2021deep, oquab2023dinov2, zhang2024tale} and have also been shown to excel in temporal correspondence matching across videos \cite{aydemir2024can, tumanyan2024dino}.
Thus, in Fig. \ref{fig: supple-feature-track}, we present additional comparisons of video tracking using the intermediate features of pretrained models, including Track4Gen, DINOv2 \cite{oquab2023dinov2}, Stable Video Diffusion \cite{blattmann2023stable}, and Zeroscope \cite{zeroscope}.
Furthermore, Fig. \ref{fig: supple-vs-dino} offers a direct comparison between Track4Gen and DINOv2 features.
While Track4Gen features demonstrate robustness, they are less effective in videos with occlusions.

\subsection{Track4Gen with DINO-Tracker}
\begin{figure*}[!htb]
    \centering
    \includegraphics[width=\textwidth]{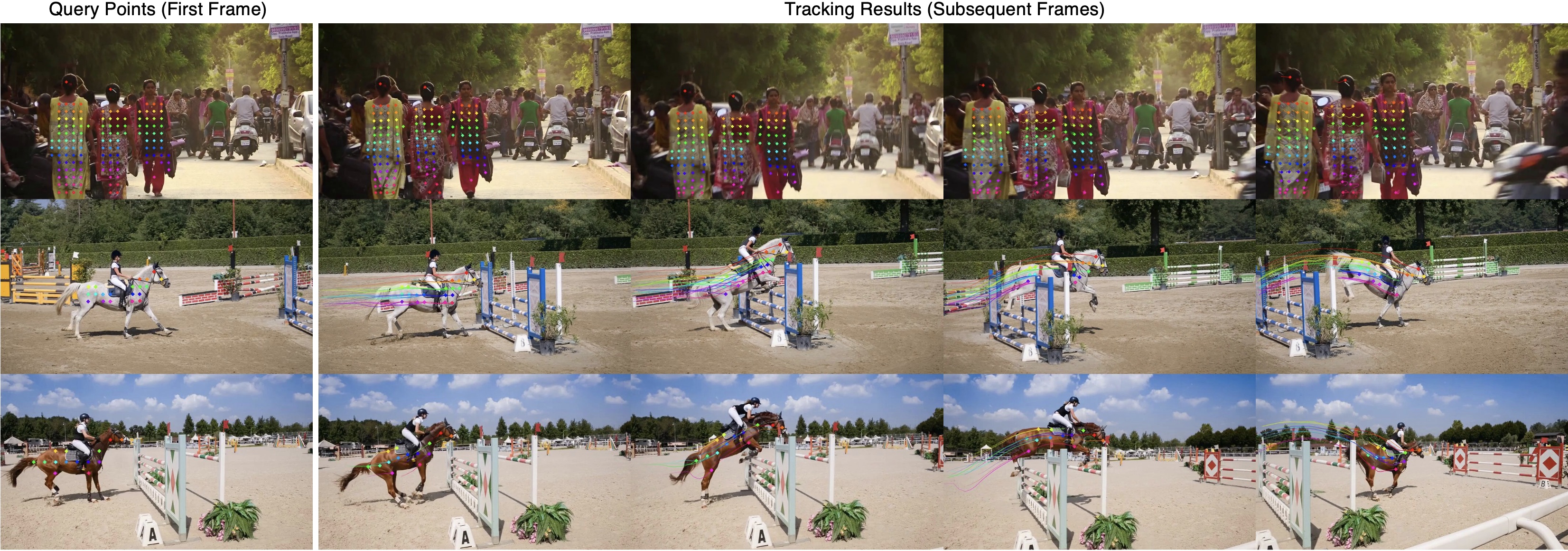}
    \caption{
    \textbf{Extending Track4Gen features with test-time adaptation \cite{tumanyan2024dino}. }
    }
    \label{fig: supple-with-dino-tracker}
\end{figure*}

\begin{figure*}[!htb]
    \centering
    \includegraphics[width=\textwidth]{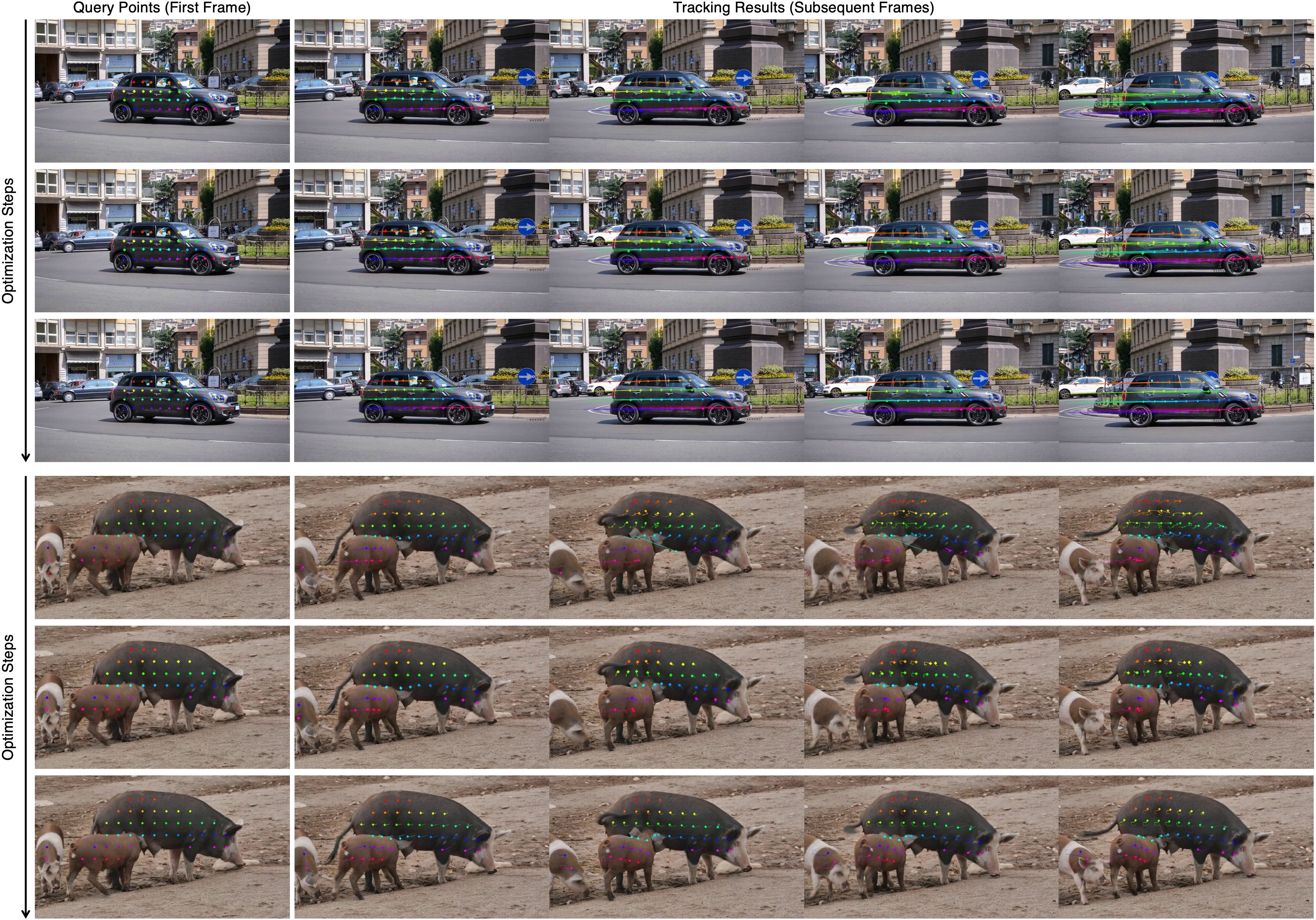}
    \caption{
    \textbf{Optimization progress visualization.}
    The first rows show tracking results using zero-shot Track4Gen features, while the third rows display results after 5,000 optimization steps.
    }
    \label{fig: supple-progress}
\end{figure*}

We present additional results of adapting Track4Gen features with DINO-Tracker \cite{tumanyan2024dino} in Fig. \ref{fig: supple-with-dino-tracker}. 
Moreover, the optimization progress is visualized in Fig. \ref{fig: supple-progress}, showing how the optical flow-guided test-time adaptation enhances the incomplete raw Track4Gen features.

\section{Discussion on Limitation and Failure}
\label{supple-sec: limitation}

\begin{figure*}[!htb]
    \centering
    \includegraphics[width=\textwidth]{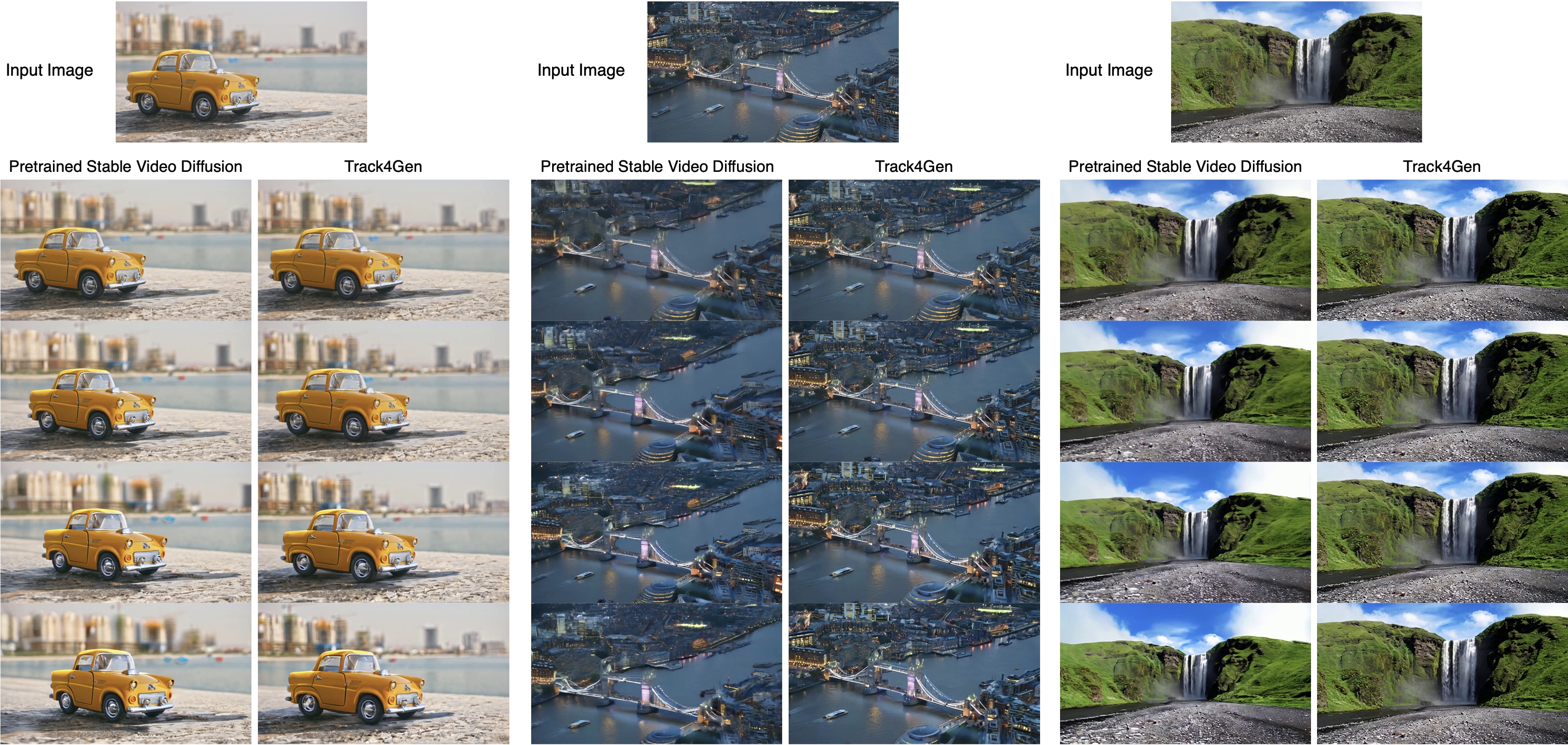}
    \caption{
    \textbf{Limitation.}
    Generated videos of Track4Gen may exhibit reduced camera motion.
    }
    \label{fig: supple-limit}
\end{figure*}

\begin{figure*}[!htb]
    \centering
    \includegraphics[width=\textwidth]{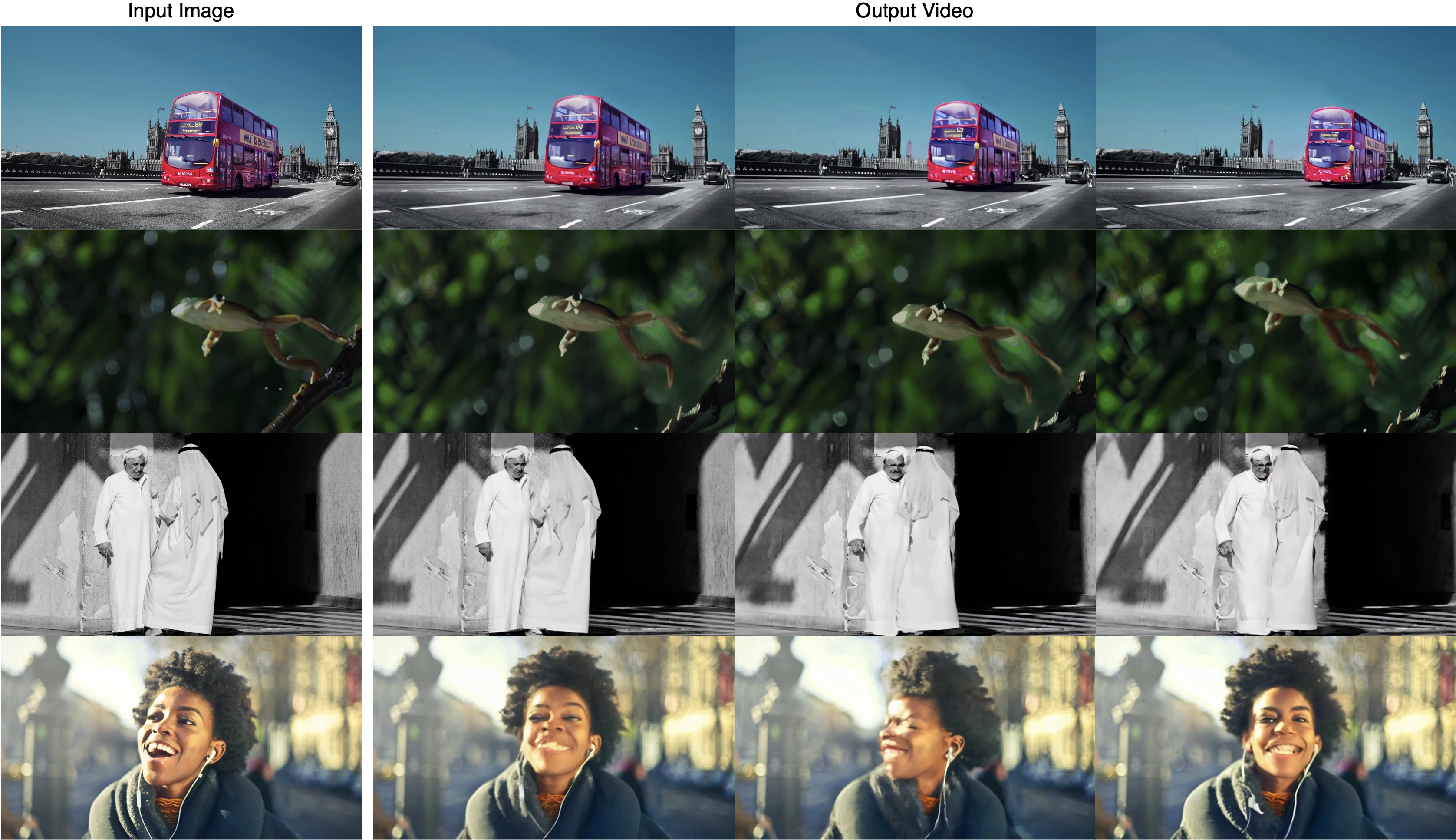}
    \caption{
    \textbf{Video generation failure cases.}
    Track4Gen may generate videos with physically unrealistic motion and artifacts on human faces. 
    For instance, the red bus (row 1) drives backward, the frog (row 2) jumps mid-air, and the faces (row 3,4) display artifacts.
    }
    \label{fig: supple-fail-gen}
\end{figure*}

For video results related to this section, please refer to \textit{page 4} of our project page.
While Track4Gen significantly enhances appearance constancy in generated videos, it tends to result in reduced camera motion compared to the original Stable Video Diffusion prior, a behavior also observed in the finetuned Stable Video Diffusion baseline. (see Fig. \ref{fig: supple-limit}).
We attribute this to the training dataset used for finetuning.
In addition, in some cases Track4Gen produces unrealistic motion and exhibit artifacts on human faces and hands, particularly when the resolution or size of the human subject in the video is small — a common limitation shared by video diffusion models \cite{kwon2024harivo}, including the baselines.
Typical failure cases of video generation are illustrated in Fig. \ref{fig: supple-fail-gen}.

\begin{figure*}[!htb]
    \centering
    \includegraphics[width=\textwidth]{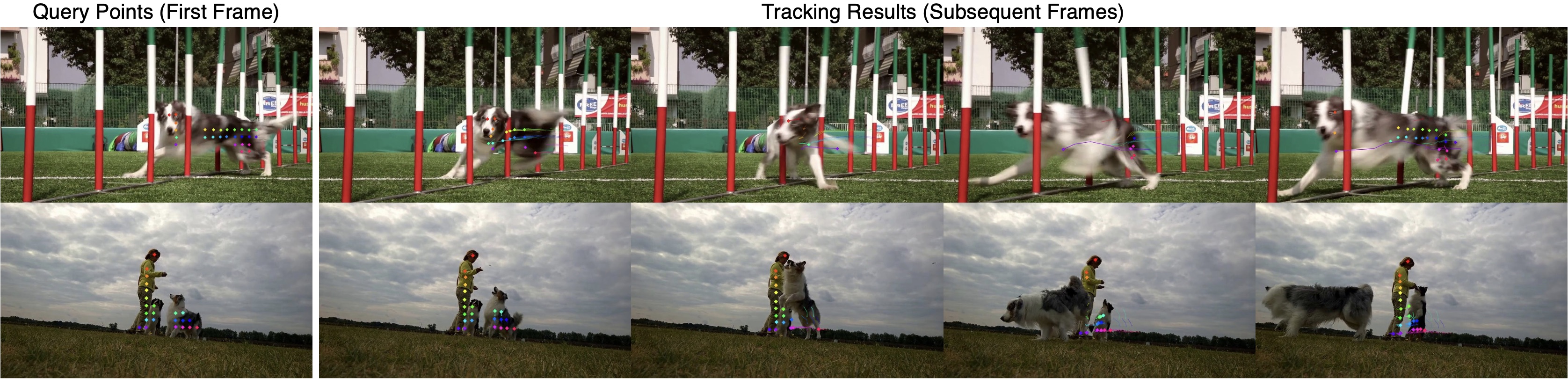}
    \caption{
    \textbf{Video tracking failure cases.}
    Track4Gen features struggle to capture point correspondences in videos with fast-moving objects or multiple semantically similar objects.
    }
    \label{fig: supple-fail-track}
\end{figure*}

We also present failure cases of real-world video tracking in Fig. \ref{fig: supple-fail-track}. 
Track4Gen features often struggle to capture accurate correspondences in videos with fast-moving objects and blurred frames. 
Additionally, Track4Gen lacks robustness in challenging videos with multiple semantically similar objects, where trajectories can shift from one object to another. An interesting direction for future work is augmenting the proposed correspondence loss with additional terms that account for occlusion predictions, which could further improve video generation performance.

\clearpage

\clearpage

\clearpage

\clearpage

\clearpage

\clearpage

\clearpage

\end{appendix}


\end{document}